\documentclass[10pt]{article}

\usepackage[a4paper,margin=1in]{geometry}
\usepackage[utf8]{inputenc}
\usepackage[T1]{fontenc}
\usepackage{amsmath,amssymb,amsfonts}
\usepackage{amsthm}
\usepackage{graphicx}
\usepackage{booktabs}
\usepackage{microtype}
\usepackage[numbers,sort&compress]{natbib}
\usepackage{enumitem}
\usepackage{float}
\usepackage[section]{placeins}
\usepackage{url}
\usepackage{hyperref}

\hypersetup{
    colorlinks=true,
    linkcolor=black,
    citecolor=black,
    urlcolor=blue
}

\setlist[itemize]{leftmargin=*,itemsep=2pt,topsep=3pt}

\theoremstyle{plain}
\newtheorem{proposition}{Proposition}

\theoremstyle{remark}

\title{
Structured Extrema Errors in Classical Surrogates for Viscous Burgers:\\
A Physics-Consistent Interpretation
}

\author{
Youssef Oubari\\
IMT Atlantique, France\\
\texttt{youssef.oubari@imt-atlantique.net}
}

\date{}

\begin{document}

\maketitle


\begin{abstract}
We study the local errors of classical machine-learning surrogate models,
which approximate the time evolution of the one-dimensional viscous Burgers
equation. Four models are compared on the
same prediction task, using the spatial grid values directly: radial basis
function (RBF) kernel ridge regression (KRR), linear Ridge, ExtraTrees, and
Random Forests. Across all four models, the one-step residual, defined here as
the true value minus the predicted value at each grid point, forms clear
curved branches near predicted maxima and minima. A more detailed analysis of
KRR shows that these errors are much more strongly related to the second
spatial derivative, which measures local curvature, than to the first spatial
derivative. Near a smooth extremum, predicted value and curvature form a local
two-branch fold. Under our local curvature-based model of the residual, this
fold predicts a leading-order near-parabolic relation between predicted value
and residual. This geometric result motivates a direct test of the Burgers advection
(transport) and diffusion (smoothing) terms. For KRR and Ridge, regression
tests on held-out trajectories, a control that breaks the spatial alignment of
the diffusion
term, and a spectral test of high-frequency content are consistent with
insufficient viscous smoothing at moderate and high viscosity. In this case,
the surrogate retains more small-scale structure than the true future state.
The same physical explanation is much weaker for the tree models. Finally, a
correction that uses only predicted quantities reduces both one-step error and
error during recursive rollout, where each prediction is used as the next
input.
\end{abstract}


\section{Introduction}
\label{sec:introduction}

Machine-learning methods are widely used to approximate solutions and
solution operators of partial differential equations (PDEs). A solution
operator maps an input field or parameter to a solution field
\citep{raissi2019pinn,lu2021deeponet,li2021fno,kovachki2023neuraloperator}.
Their accuracy is commonly reported with global measures such as mean squared
error or relative $L^2$ error
\citep{li2021fno,takamoto2022pdebench,kovachki2023neuraloperator}.
These measures summarize the prediction over the full domain, but they do not
describe where the largest errors occur or whether those errors have a
repeatable local structure. Scientific interpretability work therefore
emphasizes quantitative tests and checks based on domain knowledge
\citep{doshivelez2017rigorous,adebayo2018sanity,roscher2020explainable}.

There are also well-established links between numerical error and
differential operators. Modified-equation analysis describes discretization
errors through additional differential terms
\citep{warming1974modified,durran2010numerical}. In learned PDE models,
governing-equation residuals have also been used to measure and correct
prediction errors \citep{cao2023residual,jha2024corrector}. Here, a
governing-equation residual means the mismatch obtained when a predicted
solution is inserted into the PDE. These ideas lead to the question studied
here: can the prediction error of a learned surrogate have a simple local
structure, and can that structure be linked to terms in the governing PDE?

We study this question for the one-dimensional viscous Burgers equation, a
standard benchmark in scientific machine learning
\citep{li2021fno,takamoto2022pdebench}. The comparison uses four classical
models: radial basis function (RBF) kernel ridge regression (KRR)
\citep{scholkopf2002kernels}, linear Ridge \citep{hoerl1970ridge}, ExtraTrees
\citep{geurts2006extratrees}, and Random Forests \citep{breiman2001rf}. All
models use the same discretized spatial grid values and viscosity as input.
This gives a controlled setting for comparing their error structure.

\begin{figure}[H]
\centering
\IfFileExists{figures/rollout_krr_gain_snapshot.png}{
    \includegraphics[width=0.55\linewidth]
    {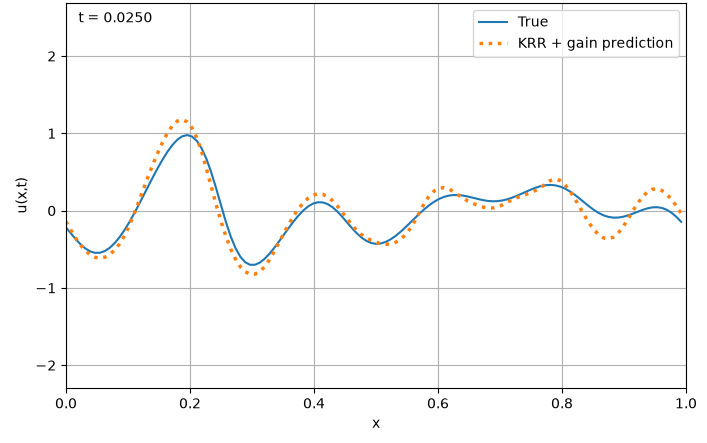}
}{
    \fbox{\parbox{0.60\linewidth}{\centering
    Insert \texttt{figures/rollout\_krr\_gain\_snapshot.png}.}}
}
\caption{
Qualitative KRR rollout example. Here, rollout means that the model is applied
repeatedly and each prediction is used as the next input. The prediction
follows the main shape of the reference solution, while some of the largest
visible errors occur near peaks and troughs. This figure is used only as
motivation. The quantitative results use the fixed protocol described in
Sec.~\ref{sec:problem_method} and Appendix~\ref{app:protocol}.
}
\label{fig:motivating_snapshot}
\end{figure}

The first observation is that the pointwise residual
$r=u_{\rm true}-\hat u$ is strongly structured, where $u_{\rm true}$ is the
true future value and $\hat u$ is the predicted value. When $r$ is plotted
against $\hat u$, clear curved branches appear near predicted maxima and
minima. The same general pattern is present for all four model families,
although its size and orientation vary between models. A qualitative KRR
example is shown in Fig.~\ref{fig:motivating_snapshot}: the main solution
shape is reproduced, while some of the largest errors occur around peaks and
troughs.

To understand why these branches appear, we first model the local discrepancy
using three simple components: amplitude change, spatial shift, and smoothing.
The model uses a translation operator, which shifts a field in space, and a
heat operator, which represents diffusion and smoothing
\citep{engel2000semigroups,pazy1983semigroups}. These components are also
consistent with standard descriptions of spatial-shift and smoothing errors
in numerical analysis \citep{warming1974modified,durran2010numerical}. To
first order, meaning that we keep only terms linear in the small changes, we
show that the residual is a combination of the predicted value, its first
derivative, and its second derivative. The derivation is given in
Appendix~\ref{app:derivations}.

This first-order model points directly to extrema. At a smooth interior
extremum, the first spatial derivative is zero by the standard first-order
condition for differentiable functions \citep{rudin1976principles}. Our
experiments show that the second-derivative term, which measures local
curvature, explains much more of the KRR residual than the first-derivative
term. This motivates a closer study of the local geometry. Using Taylor's
theorem \citep{rudin1976principles}, we show that predicted value and
curvature form a local fold: the two sides of a smooth extremum form two
branches in the value--curvature relation. When this fold is combined with
the local curvature-based residual model, it predicts a leading-order
near-parabolic relation between predicted value and residual. The full
derivation is given in Appendix~\ref{app:derivations}.

Curvature also has a direct physical role in the Burgers equation because the
viscous diffusion term is proportional to the second spatial derivative
\citep{durran2010numerical}. This connection motivates the next test. We
compare the observed residual with the Burgers advection and diffusion terms
using regression evaluated on held-out trajectories. We also use a spatial
negative control: the diffusion field is shifted so that its values are no
longer aligned with the residual at the same grid points. For KRR and Ridge,
the correctly aligned diffusion term explains a large part of the residual
near extrema, while the advection contribution is much smaller. The same
diffusion attribution is much weaker for ExtraTrees and Random Forests.

We test the diffusion interpretation in a second way. First, we define an
effective-diffusion diagnostic: under a local heat model, it measures the
amount of diffusion that would reproduce the surrogate's smoothing. Second,
we use a spectral measure of small-scale content, based on Fourier
coefficients. The heat-model relation follows from the standard heat
semigroup \citep{pazy1983semigroups}, while the spectral measure follows from
Fourier differentiation and Parseval's identity
\citep{trefethen2000spectral}. At moderate and high viscosity, both tests are
consistent with insufficient viscous smoothing for KRR and Ridge. In this
case, the surrogate retains more small-scale structure than the true future
state. The weaker tree-model attribution shows that similar residual geometry
can arise from different error mechanisms.

The final question is whether the identified structure contains useful
information. An oracle experiment, meaning a test that is allowed to use the
true residual profile, first measures how much local residual energy can be
removed. We then train a
correction that uses only quantities computed from the predicted state and
does not use the true residual at test time. The correction is tested on
held-out one-step predictions and during recursive rollout, where each
prediction is used as the next input.

The main contributions are:
\begin{itemize}
    \item We show that large residuals are concentrated near predicted extrema
    and that structured extrema-local branches occur across four classical
    surrogate families.

    \item For KRR, we show that curvature is the main first-order quantity
    associated with these residuals. A local fold analysis then explains why
    a near-parabolic structure can appear between predicted value and residual
    under the curvature-based residual model.

    \item For KRR and Ridge, regression against the Burgers terms on held-out
    trajectories, a spatial negative control, effective-diffusion estimates,
    and spectral measurements are
    consistent with insufficient viscous smoothing at moderate and high
    viscosity. The same explanation is much weaker for the tree models.

    \item A correction learned only from predicted quantities reduces KRR
    error in both one-step prediction and recursive rollout.
\end{itemize}

\section{Related Work}
\label{sec:related_work}

Machine learning has been widely used to approximate solutions and solution
operators of partial differential equations (PDEs), through physics-informed
neural networks \citep{raissi2019pinn,karniadakis2021physics} and
operator-learning methods such as DeepONet \citep{lu2021deeponet} and Fourier
neural operators \citep{li2021fno,kovachki2023neuraloperator}. The viscous
Burgers equation is a common benchmark in this literature
\citep{li2021fno,takamoto2022pdebench}. Kernel-based methods have also been
studied for learning PDE solution maps and operators
\citep{nelsen2024randomfeatures,stepaniants2023rkhs}. We use classical
raw-grid surrogates as controlled models for studying structured prediction
errors.

Our work is also related to scientific interpretability and model-error
analysis. Scientific explanations should be tested quantitatively and checked
against domain knowledge
\citep{doshivelez2017rigorous,adebayo2018sanity,roscher2020explainable}.
Modified-equation analysis connects numerical errors to differential operators
\citep{warming1974modified}, while data-driven methods aim to identify simple
dynamical models from observations \citep{brunton2016sindy}. More directly
related to our work, Cao et al.~\citep{cao2023residual} use the governing PDE
residual to correct neural-operator predictions, and
Jha~\citep{jha2024corrector} develops a residual-based correction operator for
nonlinear PDE surrogates. These works use physical structure to improve
surrogate predictions and focus mainly on estimating or correcting the error.
We start from the prediction residual $r$, study its local structure, and ask
which terms of the governing PDE explain that structure. We then test this
interpretation on held-out trajectories, with negative controls, across
different regimes, and through error correction.

\section{Problem Setting and Method}
\label{sec:problem_method}

\subsection{Burgers equation and prediction task}

We study the standard one-dimensional viscous Burgers equation
\citep{takamoto2022pdebench,durran2010numerical} on the periodic domain
$x\in[0,1)$:
\begin{equation}
\partial_t u + u\,\partial_x u
=
\nu\,\partial_{xx}u,
\label{eq:burgers}
\end{equation}
where $u(x,t)$ is the solution field and $\nu>0$ is the viscosity.

For a fixed prediction horizon $H$, we define the numerical flow map
$\Phi_H$ by
\begin{equation}
u^{n+1}
=
\Phi_H(u^n,\nu),
\qquad
\Phi_H:
\mathbb{R}^{N}\times\mathbb{R}_{+}
\rightarrow
\mathbb{R}^{N},
\label{eq:true_flow}
\end{equation}
where $u^n\in\mathbb{R}^N$ is the reference solution sampled on the spatial
grid at the current stored time.

We define a raw-grid surrogate as a learned function
\begin{equation}
\hat u^{n+1}
=
f(u^n,\nu),
\qquad
f:
\mathbb{R}^{N+1}
\rightarrow
\mathbb{R}^{N},
\label{eq:surrogate_map}
\end{equation}
where $\hat u^{n+1}$ is the predicted next state. In one-step evaluation,
every prediction starts from a true input state $u^n$.

\subsection{Data and surrogate models}

We generate the initial conditions using the following five-mode family:
\begin{equation}
u(x,0)
=
\sum_{k=1}^{5}
a_k\sin(2\pi kx+\phi_k),
\qquad
a_k\sim\mathcal U[-1,1],
\quad
\phi_k\sim\mathcal U[0,2\pi].
\label{eq:initial_conditions}
\end{equation}
Equation~\eqref{eq:initial_conditions} is an experimental choice made in this
work to generate smooth periodic initial states with different shapes.

For each trajectory, we sample one viscosity from
\[
\nu\in\{0.01,0.02,0.05,0.1\}.
\]
This is an experimental choice in our data-generation protocol.

The spatial grid has $N=128$ points. We generate the reference trajectories
with a fourth-order Runge--Kutta (RK4) time integrator, a standard explicit
time-integration method \citep{durran2010numerical}. We use time step
$\Delta t=10^{-4}$ for $5000$ RK4 steps and store a state every $250$ solver
steps. We therefore define the canonical one-step horizon as
\[
H=250\Delta t=0.025.
\]

The canonical data set contains $30$ trajectories. We split them into $24$
training and $6$ validation trajectories using seed $42$. Full numerical and
model settings are listed in Appendix~\ref{app:protocol}.

We compare RBF KRR \citep{scholkopf2002kernels}, linear Ridge regression
\citep{hoerl1970ridge}, ExtraTrees \citep{geurts2006extratrees}, and Random
Forests \citep{breiman2001rf}. KRR is the strongest model in the canonical
comparison and gives the clearest local branches, so we use it as the
reference model for the detailed geometric, correction, and rollout analyses.

\subsection{Prediction residual and predicted extrema}

For a true one-step target $u_{\rm true}$ and its surrogate prediction
$\hat u$, we define the pointwise residual by
\begin{equation}
r
=
u_{\rm true}-\hat u.
\label{eq:residual}
\end{equation}

We approximate the first and second spatial derivatives with standard centered
periodic finite differences \citep{durran2010numerical}:
\begin{equation}
\partial_x\hat u_i
\simeq
\frac{\hat u_{i+1}-\hat u_{i-1}}
{2\Delta x},
\qquad
\partial_{xx}\hat u_i
\simeq
\frac{\hat u_{i+1}-2\hat u_i+\hat u_{i-1}}
{\Delta x^2}.
\label{eq:finite_differences}
\end{equation}

We define a predicted local maximum as a grid point $i$ satisfying
\[
\hat u_i>\hat u_{i-1},
\qquad
\hat u_i>\hat u_{i+1},
\]
and define a predicted local minimum by reversing both inequalities.

We define the \emph{near-extrema} mask as all grid points within periodic
radius $3$ of a predicted extremum.

For local geometry, we define a radius-$5$ window around each predicted
extremum, giving $11$ grid points per window. We define active windows using
the $90$th-percentile rule stated precisely in
Appendix~\ref{app:protocol}.

\subsection{Local geometry}

We use standard principal-component analysis (PCA) to summarize the local
shape and orientation of each residual branch \citep{jolliffe2002pca}.
For each active local window, we define the centered vectors
\begin{equation}
p_j=
\begin{bmatrix}
\hat u_j-\bar u\\
r_j-\bar r
\end{bmatrix}.
\label{eq:centered_points}
\end{equation}

Following the standard covariance formulation of PCA
\citep{jolliffe2002pca}, we define
\begin{equation}
C
=
\frac{1}{m}
\sum_{j=1}^{m}
p_jp_j^\top.
\label{eq:local_covariance}
\end{equation}

Let $\lambda_1\ge\lambda_2$ be the eigenvalues of $C$ and let $v_1$ be the
principal eigenvector. Following the standard PCA interpretation of the first
principal direction and eigenvalue spread \citep{jolliffe2002pca}, we define
the branch angle $\theta$ and elongation ratio $\rho$ as
\begin{equation}
\theta
=
\arctan\!\left(
\frac{|(v_1)_2|}
{|(v_1)_1|}
\right),
\qquad
\rho
=
\sqrt{\frac{\lambda_1}{\lambda_2}}.
\label{eq:orientation}
\end{equation}
Thus $\theta=0^\circ$ corresponds to a branch aligned with the predicted-value
axis and $\theta=90^\circ$ to a branch aligned with the residual axis $r$.
The angle and ratio in Eq.~\eqref{eq:orientation} are summary measures defined
in this work from the standard PCA quantities.

For descriptive conic fits, we use the standard general quadratic form for a
planar conic \citep{fitzgibbon1999ellipse}:
\begin{equation}
aX^2+bXY+cY^2+dX+eY+g=0,
\label{eq:empirical_conic}
\end{equation}
where we standardize $X=\hat u$ and $Y=r$ inside each radius-$5$ window.

We estimate the coefficients algebraically using the right singular vector
associated with the smallest singular value of the corresponding design
matrix, following standard algebraic conic-fitting methods
\citep{fitzgibbon1999ellipse}.

The quantity $b^2-4ac$ is the standard quadratic-part discriminant used
in conic classification \citep{fitzgibbon1999ellipse}. We use it only to
construct descriptive labels for our fitted windows. For numerical stability,
we define the normalized discriminant
\begin{equation}
\Delta
=
\frac{b^2-4ac}
{a^2+b^2+c^2}.
\label{eq:conic_discriminant}
\end{equation}

We also define the numerical tolerance
\[
\tau=10^{-3}.
\]
We call a fit ellipse-like if $\Delta<-\tau$, hyperbola-like if
$\Delta>\tau$, and parabola-like if $|\Delta|\le\tau$. The normalization and
the value of $\tau$ are choices made in this work for numerical
classification.

\subsection{Evaluation protocol}

For $M$ evaluated states on an $N$-point grid, we define the pointwise
mean squared error by
\begin{equation}
\operatorname{MSE}
=
\frac{1}{MN}
\sum_{i=1}^{M}
\|u_i-\hat u_i\|_2^2.
\label{eq:mse}
\end{equation}

We define the relative $L^2$ error over the same set of states by
\begin{equation}
\operatorname{RelL2}
=
\frac{
\left(\sum_{i=1}^{M}\|u_i-\hat u_i\|_2^2\right)^{1/2}
}{
\left(\sum_{i=1}^{M}\|u_i\|_2^2\right)^{1/2}
}.
\label{eq:rell2}
\end{equation}
Both equations are definitions of the evaluation measures used in this work.

For the physics attribution in Sec.~\ref{sec:physics_attribution}, we use
five-fold cross-fitting by trajectory over all $30$ trajectories. Therefore
every residual used in that analysis comes from a surrogate that was not
trained on the same trajectory. Bootstrap intervals are computed by resampling
whole trajectories rather than individual grid points.

\section{Geometric Analysis}
\label{sec:geometric_analysis}

\subsection{A first-order residual model}

We introduce a local model that separates three simple differences between the
reference solution and the surrogate prediction: amplitude, spatial shift, and
smoothing.

For a periodic field $v$, we use the standard translation operator
\citep{engel2000semigroups}, written as
\[
T_\delta v(x)=v(x+\delta).
\]

We also use the standard heat semigroup
$e^{\beta\partial_{xx}}$ \citep{pazy1983semigroups}. Motivated by these two
operators and by the standard separation of phase and dissipative numerical
errors \citep{warming1974modified,durran2010numerical}, we introduce the local
error model
\begin{equation}
u_{\rm true}
\approx
(1+\alpha)
e^{\beta\partial_{xx}}
T_\delta\hat u.
\label{eq:operator_model}
\end{equation}

Equation~\eqref{eq:operator_model} is a model introduced in this work. It is
used only as a local description of possible amplitude, shift, and smoothing
errors.

Using the standard first-order expansions of the translation and heat
operators, we show in Appendix~\ref{app:derivations} that
Eq.~\eqref{eq:operator_model} gives
\begin{equation}
r
\approx
\alpha\hat u
+
\delta\,\partial_x\hat u
+
\beta\,\partial_{xx}\hat u
\label{eq:first_order_residual}
\end{equation}
to first order.

For a differentiable continuous field, the first derivative is zero at an
interior local extremum by the standard first-order condition
\citep{rudin1976principles}. Our extrema are detected on a discrete grid, so
we use
\[
\partial_x\hat u\simeq0
\]
as a local approximation near the detected extrema.

Using this approximation in Eq.~\eqref{eq:first_order_residual}, we obtain
\begin{equation}
r
\approx
\alpha\hat u
+
\beta\partial_{xx}\hat u.
\label{eq:extremum_residual}
\end{equation}

This relation motivates testing curvature near predicted extrema.

\subsection{Fold geometry near an extremum}

For the local analysis, we consider a sufficiently smooth local
representation of the predicted field. Let $x_0$ be a non-degenerate
continuous extremum. We define
\[
U_0=\hat u(x_0),
\qquad
c_2=\partial_{xx}\hat u(x_0)\neq0,
\qquad
c_3=\partial_{xxx}\hat u(x_0)\neq0.
\]

For a nearby point $x$, we define
\[
X=\hat u(x),
\qquad
Z=\partial_{xx}\hat u(x).
\]

\begin{proposition}[Local fold relation]
\label{prop:fold}
Under the smoothness and non-degeneracy assumptions above, we show in
Appendix~\ref{app:derivations} that, to leading order,
\begin{equation}
(Z-c_2)^2
\approx
\frac{2c_3^2}{c_2}
(X-U_0).
\label{eq:fold_relation}
\end{equation}
\end{proposition}

Equation~\eqref{eq:fold_relation} is locally parabolic in the $(X,Z)$ plane.

Combining Eq.~\eqref{eq:fold_relation} with the first-order extremum relation
Eq.~\eqref{eq:extremum_residual}, we show in
Appendix~\ref{app:derivations} that, for $\beta\neq0$,
\begin{equation}
\boxed{
\left(
r-\alpha X-\beta c_2
\right)^2
\approx
\beta^2
\frac{2c_3^2}{c_2}
(X-U_0).
}
\label{eq:residual_conic}
\end{equation}

Equation~\eqref{eq:residual_conic} gives a leading-order quadratic relation
in the $(X,r)$ plane under the local curvature-based residual model. Its
leading-order class is parabolic. Finite-window empirical fits can also appear
ellipse-like or hyperbola-like because they include higher-order terms,
finite-window effects, discretization, and fitting error.

The leading-order result describes the local geometric structure; it does not
imply that a conic fitted to a finite window will always recover the correct
local shape. Figure~\ref{fig:representative_conics} shows representative
empirical windows and their fitted conics. Some fits follow the observed
branch well, while others select a different conic type from the shape
suggested by the data. We study these finite-window fitting limitations in
Appendix~\ref{app:geometry_details}.

\section{Experiments and Interpretation}
\label{sec:experiments}

\subsection{Structured residuals appear across model families}
\label{sec:crossmodel}

\begin{figure}[H]
\centering
\IfFileExists{figures_interpretability/residual_comparison_common_scale.png}{
  \includegraphics[width=0.96\linewidth]
  {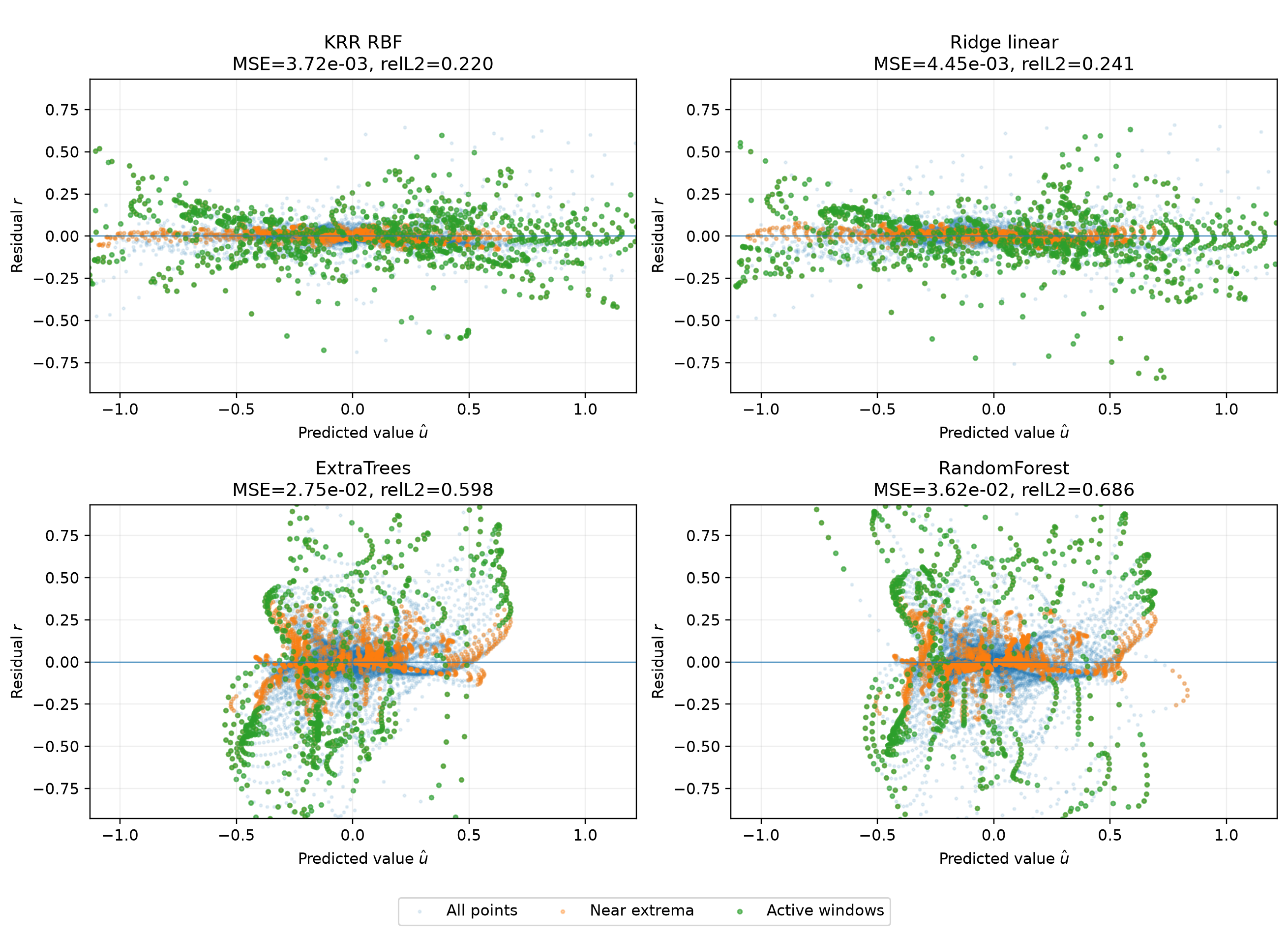}
}{
  \IfFileExists{residual_comparison_common_scale.png}{
    \includegraphics[width=0.96\linewidth]
    {residual_comparison_common_scale.png}
  }{
    \fbox{\parbox{0.88\linewidth}{\centering
    Insert \texttt{residual\_comparison\_common\_scale.png}.}}
  }
}
\caption{
Residual geometry across four classical surrogate families on the same
canonical validation set, shown with common axes. Structured off-axis
branches are visible across all models. KRR and Ridge remain comparatively
concentrated around $r=0$, while ExtraTrees and Random Forests show larger
residual magnitudes and branches oriented more strongly along the $r$-axis.
}
\label{fig:crossmodel}
\end{figure}

Figure~\ref{fig:crossmodel} shows the main empirical observation on the same
canonical validation set and with common axes. In all four model families,
$r=u_{\rm true}-\hat u$ forms coherent off-axis branches. KRR and Ridge remain
comparatively concentrated around $r=0$. ExtraTrees and Random Forests show
larger residual magnitudes, with branches oriented more strongly along the
$r$-axis.

Table~\ref{tab:crossmodel} reports the error and the local geometry measures
defined in Sec.~\ref{sec:problem_method}. The tree branches are close to the
$r$-axis, while KRR and Ridge have smaller orientation angles. The number of
active branches is also stable for the stochastic tree models across three
seeds: $67.3\pm2.1$ for ExtraTrees and $66.7\pm1.5$ for Random Forests.

\begin{table}[H]
\centering
\small
\caption{Canonical cross-model error and active-window geometry.}
\label{tab:crossmodel}
\begin{tabular}{lccccc}
\toprule
Model & MSE & Rel.\ $L^2$ & Active & Median $\theta$ & Median $\rho$ \\
\midrule
RBF KRR
& $3.72\times10^{-3}$
& 0.220
& 93
& $49.7^\circ$
& 2.51
\\

Linear Ridge
& $4.45\times10^{-3}$
& 0.241
& 103
& $48.8^\circ$
& 2.58
\\

ExtraTrees
& $2.75\times10^{-2}$
& 0.598
& 68
& $85.9^\circ$
& 8.29
\\

Random Forest
& $3.62\times10^{-2}$
& 0.686
& 67
& $83.0^\circ$
& 7.25
\\
\bottomrule
\end{tabular}
\end{table}

The common result is that classical raw-grid surrogates can develop a
structured extrema-local error. Its scale and orientation depend strongly on
the model family.

\subsection{The largest residuals are concentrated near extrema}
\label{sec:extrema}

\begin{figure}[H]
\centering
\IfFileExists{
figures_part1_conditions/part1_plain_krr_one_step_residual_arcs_near_extrema.png
}{
    \includegraphics[width=0.72\linewidth]
    {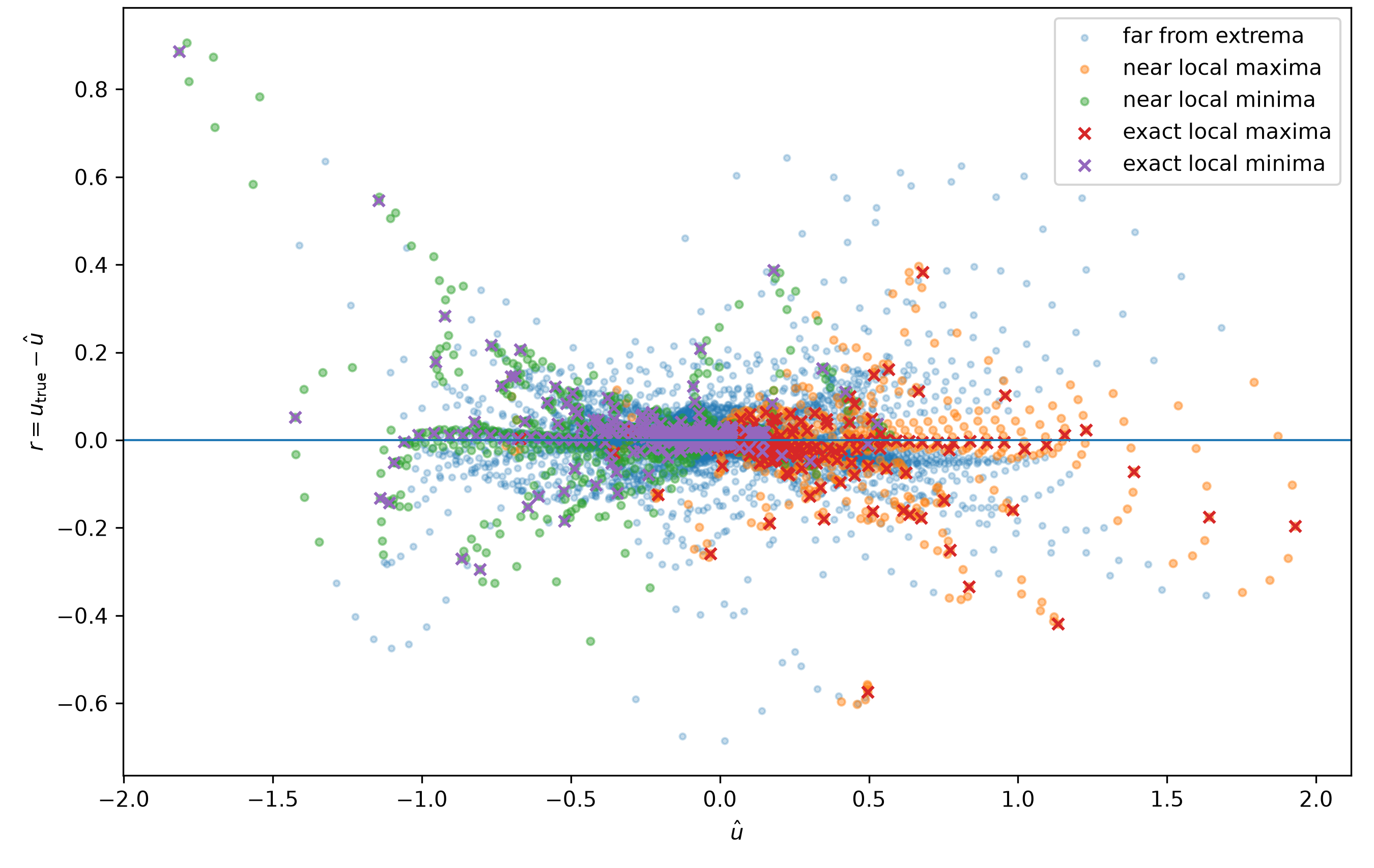}
}{
    \fbox{\parbox{0.66\linewidth}{\centering
    Insert
    \texttt{part1\_plain\_krr\_one\_step\_residual\_arcs\_near\_extrema.png}.}}
}
\caption{
Canonical one-step KRR residuals. Points within the radius-$3$ neighborhood
of predicted extrema trace the main off-axis residual branches.
}
\label{fig:extrema_arcs}
\end{figure}

We next focus on KRR. Figure~\ref{fig:extrema_arcs} highlights points inside
the radius-$3$ neighborhood of predicted extrema. These points trace the
off-axis branches and show directly that the largest residual structures are
tied to peaks and troughs.

The near-extrema mask contains $23.9\%$ of all grid points, but it contains
$39.6\%$ of the largest $5\%$ of residuals and $40.9\%$ of the largest
$1\%$. Table~\ref{tab:extrema} reports the corresponding residual statistics.

\begin{table}[H]
\centering
\small
\caption{KRR residual concentration in the canonical protocol.}
\label{tab:extrema}
\begin{tabular}{lcccc}
\toprule
Subset
& $N$
& Mean $|r|$
& 95th pct.\ $|r|$
& Max $|r|$
\\
\midrule

All points
& 15,360
& 0.027
& 0.110
& 0.907
\\

Exact extrema
& 552
& 0.035
& 0.155
& 0.886
\\

Near extrema
& 3,668
& 0.036
& 0.158
& 0.907
\\

Far from extrema
& 11,692
& 0.024
& 0.094
& 0.686
\\
\bottomrule
\end{tabular}
\end{table}

These results show that extrema occupy a minority of grid points but contain a
disproportionate share of the largest residuals.

\subsection{Testing the first-order residual model}
\label{sec:first_order_test}

To test the three quantities predicted by
Eq.~\eqref{eq:first_order_residual}, we use a separate high-sample diagnostic
protocol. This experiment is distinct from the canonical protocol used in
Secs.~\ref{sec:crossmodel} and~\ref{sec:extrema}. It trains KRR on $5{,}000$
subsampled training states and evaluates $20{,}000$ validation states, giving
$2{,}560{,}000$ pointwise residual samples. In this diagnostic setting, the
selected KRR parameters are $\alpha_{\rm KRR}=10^{-4}$ and
$\gamma_{\rm RBF}=10^{-4}$, and the one-step MSE is
$7.8596\times10^{-6}$. The saved-time horizon and sampling differ from the
canonical experiment, so we do not compare its absolute MSE with the canonical
MSE. The full protocol is summarized in Appendix~\ref{app:protocol}.

On this diagnostic set, we define the linear regression
\begin{equation}
r
=
a_0
+
a_1\hat u
+
a_2\partial_x\hat u
+
a_3\partial_{xx}\hat u
+
a_4\nu
+
\varepsilon.
\label{eq:diagnostic_regression}
\end{equation}

Equation~\eqref{eq:diagnostic_regression} is introduced in this work only to
compare how strongly the first-order quantities are related to the observed
residual.

Table~\ref{tab:diagnostic_regression} shows three clear patterns. First,
$\partial_x\hat u$ explains almost none of the residual variance. Second,
curvature is the dominant single term. Third, the full model explains more
variance at extrema than away from them, especially at maxima.

\begin{table}[H]
\centering
\small
\caption{
KRR diagnostic regression on the large diagnostic protocol. The
viscosity-only contribution is negligible and is omitted from the table.
}
\label{tab:diagnostic_regression}
\begin{tabular}{lcccc}
\toprule
Subset
& $R^2_{\rm full}$
& $R^2(\hat u)$
& $R^2(\partial_x\hat u)$
& $R^2(\partial_{xx}\hat u)$
\\
\midrule

All points
& 0.091
& 0.035
& 0.001
& 0.088
\\

Maxima
& 0.274
& 0.057
& 0.003
& 0.242
\\

Minima
& 0.173
& 0.046
& 0.002
& 0.156
\\

All extrema
& 0.225
& 0.052
& 0.002
& 0.185
\\

Non-extrema
& 0.087
& 0.034
& 0.001
& 0.084
\\
\bottomrule
\end{tabular}
\end{table}

The first-derivative term has very little explanatory power in this diagnostic
experiment. The curvature term has the largest explanatory power among the
three first-order quantities, consistent with the reduced extremum relation in
Eq.~\eqref{eq:extremum_residual}.

\begin{figure}[H]
\centering
\IfFileExists{
figures_part1_conditions/stepC_u_hat_vs_dxx_u_hat_near_extrema.png
}{
    \includegraphics[width=0.66\linewidth]
    {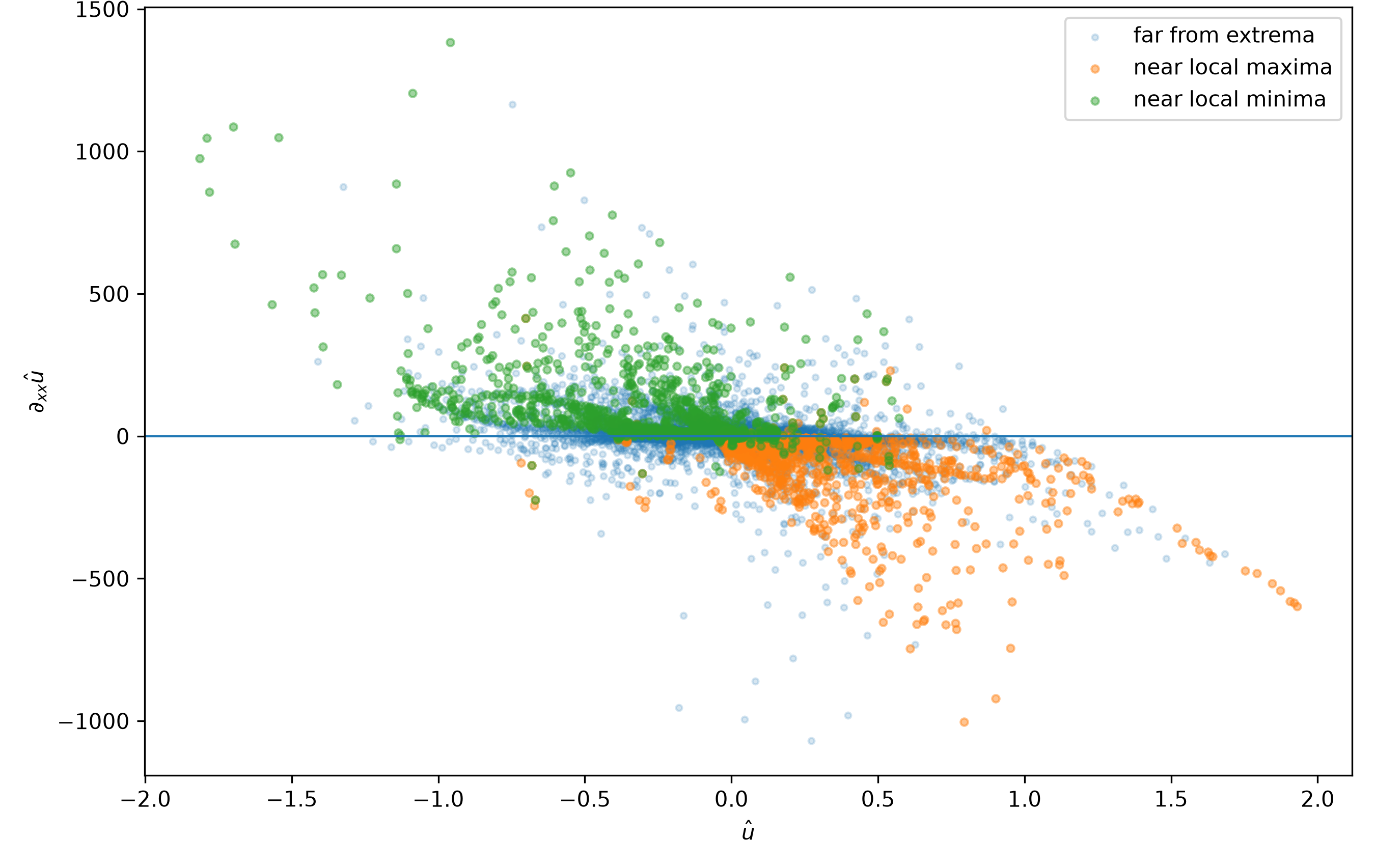}
}{
    \fbox{\parbox{0.60\linewidth}{\centering
    Insert
    \texttt{stepC\_u\_hat\_vs\_dxx\_u\_hat\_near\_extrema.png}.}}
}
\caption{
Predicted value versus predicted curvature near KRR extrema. The structured
value--curvature relation is consistent with the fold geometry shown in
Sec.~\ref{sec:geometric_analysis}.
}
\label{fig:value_curvature}
\end{figure}

We now return to the canonical protocol for the geometric plots and the
remaining KRR analyses. Figure~\ref{fig:value_curvature} shows the relation between predicted value
and predicted curvature near extrema. The observed fold is consistent with
the relation that we show in Eq.~\eqref{eq:fold_relation}.

Detailed finite-window conic counts and held-out conic fits are reported in
Appendix~\ref{app:geometry_details}.

\subsection{Connecting the residual to the Burgers equation}
\label{sec:physics_attribution}

The Burgers equation in Eq.~\eqref{eq:burgers} contains separate advection and
diffusion terms. Evaluating these terms on the predicted field, we define
\begin{equation}
A(\hat u)
=
-\hat u\,\partial_x\hat u,
\qquad
D(\hat u,\nu)
=
\nu\,\partial_{xx}\hat u.
\label{eq:burgers_fields}
\end{equation}

To test how strongly these two terms are related to the observed residual, we
define the diagnostic regression
\begin{equation}
\boxed{
\frac{r}{H}
=
b_0
+
a\hat u
+
\eta A(\hat u)
+
\gamma D(\hat u,\nu)
+
\varepsilon.
}
\label{eq:physics_regression}
\end{equation}

Equation~\eqref{eq:physics_regression} is an attribution model introduced in
this work. In Appendix~\ref{app:derivations}, we show from the fundamental
theorem of calculus and Eq.~\eqref{eq:burgers} why the residual rate $r/H$
can be compared with the time-averaged Burgers terms.

For a feature $q$, we define its held-out contribution by
\begin{equation}
\Delta R_q^2
=
R_{\rm full}^2
-
R_{\rm without\ q}^2.
\label{eq:drop_one}
\end{equation}

We compute all values with trajectory-level cross-fitting.

As a spatial negative control, we periodically shift the diffusion field
inside each state. This keeps the same diffusion values but breaks their
pointwise alignment with the residual. We define the shift gap as the
difference between the held-out $R^2$ obtained with the correctly aligned
diffusion field and the held-out $R^2$ obtained with the shifted field.

Table~\ref{tab:physics_attribution} shows a large diffusion contribution and
a large spatial shift gap for KRR and Ridge, while the advection contribution
is much smaller. ExtraTrees and Random Forests show much smaller diffusion
contributions and shift gaps.

\begin{table}[H]
\centering
\small
\caption{
Held-out attribution near predicted extrema. ``Shift gap'' is the loss of
held-out $R^2$ after spatially shifting the diffusion field.
}
\label{tab:physics_attribution}
\begin{tabular}{lcccc}
\toprule
Model
& Full $R^2$
& $\Delta R^2_{\rm diff}$
& $\Delta R^2_{\rm adv}$
& Shift gap
\\
\midrule

RBF KRR
& 0.420
& 0.386
& 0.032
& 0.392
\\

Linear Ridge
& 0.534
& 0.407
& 0.051
& 0.438
\\

ExtraTrees
& 0.101
& 0.010
& 0.001
& 0.010
\\

Random Forest
& 0.018
& 0.034
& $<0.001$
& 0.035
\\
\bottomrule
\end{tabular}
\end{table}

For Random Forest, $\Delta R^2_{\rm diff}$ is larger than the full-model
$R^2$. The reduced held-out model obtained after removing diffusion has
negative $R^2$. By the definition $R^2=1-\mathrm{SSE}/\mathrm{SST}$, this
happens when the held-out squared error is larger than the total held-out
variance.

Because the diffusion feature $D(\hat u,\nu)$ defined in
Eq.~\eqref{eq:burgers_fields} is proportional to curvature, this result gives
a physical interpretation of the curvature result. The spatial negative
control and the spectral test below provide additional evidence.

\subsection{Effective diffusion and spectral roughness}
\label{sec:effective_diffusion}

To interpret the sign and scale of the fitted diffusion coefficient $\gamma$,
we introduce the following short-horizon local comparison:
\begin{equation}
u_{\rm true}
\approx
e^{H\nu\partial_{xx}}v,
\qquad
u_{\rm sur}
\approx
e^{H\nu_{\rm eff}\partial_{xx}}v.
\label{eq:effective_diffusion_model}
\end{equation}

Equation~\eqref{eq:effective_diffusion_model} uses the standard heat semigroup
\citep{pazy1983semigroups}. It is a diagnostic model introduced in this work.
The parameter $\nu_{\rm eff}$ is used only as a local diagnostic quantity.

Using the standard first-order expansion of the heat semigroup, we show in
Appendix~\ref{app:derivations} that, under this local model,
\begin{equation}
\boxed{
\frac{\nu_{\rm eff}}{\nu}
\approx
1-\gamma.
}
\label{eq:effective_diffusion_ratio}
\end{equation}

We focus the interpretation on the moderate- and high-viscosity regimes
$\nu=0.05$ and $\nu=0.10$. Table~\ref{tab:effective_sweep} reports the
resulting effective-diffusion ratios at short and long horizons.

We also define a spectral diagnostic that does not use the residual
regression. Let $\widetilde u_k$ denote the Fourier coefficients of $u$.
Using Fourier differentiation and Parseval's identity
\citep{trefethen2000spectral}, we show in
Appendix~\ref{app:derivations} that second-derivative energy weights frequency
$k$ proportionally to $k^4$. Based on this result, we define
\begin{equation}
\mathcal R(u)
=
\frac{
\sum_k
k^4
|\widetilde u_k|^2
}{
\sum_k
|\widetilde u_k|^2
},
\qquad
Q_{\mathcal R}
=
\frac{
\mathcal R(\hat u)
}{
\mathcal R(u_{\rm true})
}.
\label{eq:roughness}
\end{equation}

By definition, $Q_{\mathcal R}>1$ means that the prediction has more
high-frequency content relative to its total spectral energy than the
corresponding true future state.

\begin{table}[H]
\centering
\small
\caption{
Effective-diffusion and spectral-roughness diagnostics in the
moderate/high-viscosity regimes.
}
\label{tab:effective_sweep}
\begin{tabular}{llcccc}
\toprule
& &
\multicolumn{2}{c}{$H=0.025$}
&
\multicolumn{2}{c}{$H=0.20$}
\\
\cmidrule(lr){3-4}
\cmidrule(lr){5-6}

Model
& $\nu$
& $\nu_{\rm eff}/\nu$
& $Q_{\mathcal R}$
& $\nu_{\rm eff}/\nu$
& $Q_{\mathcal R}$
\\
\midrule

KRR
& 0.05
& 0.728
& 2.40
& 0.920
& 5.66
\\

KRR
& 0.10
& 0.750
& 2.90
& 0.965
& 6.97
\\

Ridge
& 0.05
& 0.681
& 3.24
& 0.911
& 6.64
\\

Ridge
& 0.10
& 0.712
& 4.81
& 0.958
& 7.24
\\
\bottomrule
\end{tabular}
\end{table}

At $H=0.025$, all four moderate/high-viscosity values satisfy
$\nu_{\rm eff}/\nu<1$, and all four roughness ratios satisfy
$Q_{\mathcal R}>1$. Together, these results are consistent with insufficient
viscous smoothing: the surrogate retains more small-scale structure than the
true future state.

The two diagnostics have different dependence on the prediction horizon. In
our experiments, $\nu_{\rm eff}/\nu$ moves toward one as $H$ increases, while
the roughness ratio grows. We do not assign a specific cause to this
difference. The complete viscosity--horizon sweep is given in
Appendix~\ref{app:physics_sweep}.

\subsection{Can the identified structure help reduce the error?}
\label{sec:correction}

A useful interpretation should explain a substantial part of the error and
should provide information that can be used on unseen trajectories. We test
both levels.

\paragraph{Oracle geometric upper bound.}

We first define an oracle test in which a local amplitude--curvature residual
profile is fitted using the true residual from the same active KRR window.
Because the residual being corrected is used during fitting, this test
measures an upper bound on how much of the local structured error can be
removed.

The oracle fit removes a median $85\%$ of active-window residual energy, and
$97\%$ of active windows have more than half of their residual energy removed.

\begin{figure}[H]
\centering
\IfFileExists{
figures_correction/stepH_oracle_conic_upper_error_graphs.png
}{
    \includegraphics[width=0.70\linewidth]
    {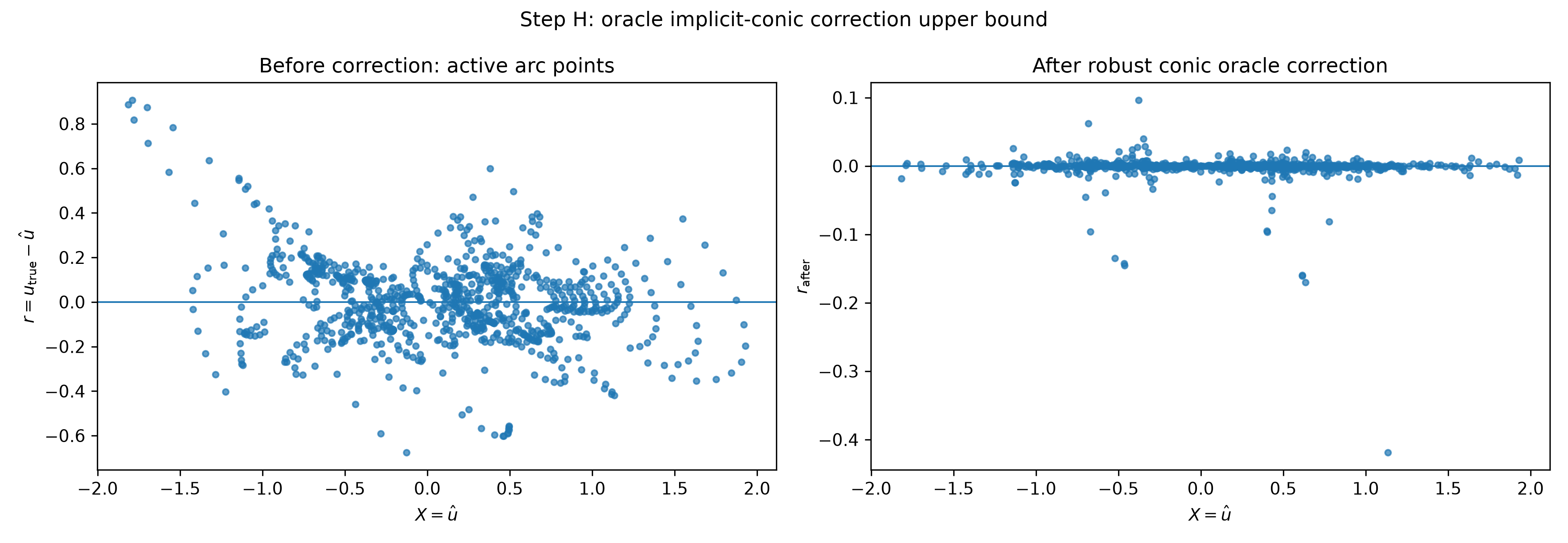}
}{
    \fbox{\parbox{0.64\linewidth}{\centering
    Insert
    \texttt{stepH\_oracle\_conic\_upper\_error\_graphs.png}.}}
}
\caption{
Oracle geometric upper bound on active KRR windows. The local correction is
fitted using each window's true residual, so this experiment measures
achievable local error removal when the residual profile is known.
}
\label{fig:oracle}
\end{figure}

\paragraph{Fully non-oracle correction.}

We next define a correction that uses only quantities available from the
prediction. The correction uses radius-$5$ profiles of
$\hat u$, $\partial_x\hat u$, and $\partial_{xx}\hat u$, local scalar
summaries, and $\nu$, with separate models for maxima and minima. The main
profile predictor uses gradient boosting \citep{friedman2001gbm} together with
a frequency-band calibration. All correction strengths are selected using
training trajectories only; no validation residual is used before evaluation.
The full procedure is defined in Appendix~\ref{app:correction_details}.

The profile--spectral correction reduces all-point MSE by $31.24\%$,
near-extrema MSE by $49.54\%$, and active-window MSE by $45.22\%$.
Relative $L^2$ decreases from $0.220$ to $0.183$.

We define the fraction of active-window residual energy removed, denoted by
$E$, in Appendix~\ref{app:correction_details}. For the main correction, its
median value is
\[
E=0.534.
\]

\begin{table}[H]
\centering
\small
\caption{
Main non-oracle one-step KRR correction result, canonical protocol.
Reductions are relative to plain KRR.
}
\label{tab:correction_main}
\begin{tabular}{lccccc}
\toprule
Method
& All red.
& Near red.
& Active red.
& Rel.\ $L^2$
& Median $E$
\\
\midrule

Profile + spectral filter
& 31.24\%
& 49.54\%
& 45.22\%
& 0.183
& 0.534
\\
\bottomrule
\end{tabular}
\end{table}

The residual plot confirms that the correction acts directly on the structured
branches. Small and moderate branches move toward $r=0$, while the largest
branches remain only partly corrected.

\begin{figure}[H]
\centering
\IfFileExists{
figures_correction/canonical_correction_before_after.png
}{
    \includegraphics[width=0.72\linewidth]
    {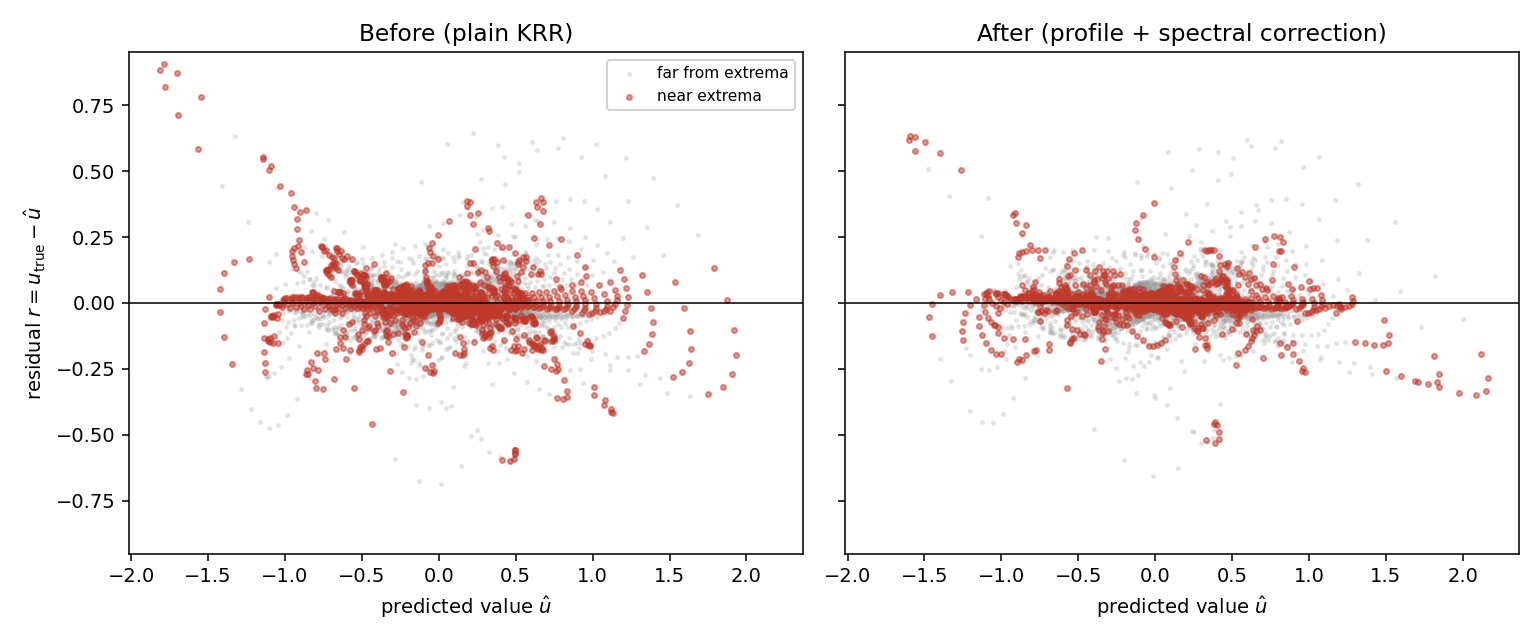}
}{
    \fbox{\parbox{0.66\linewidth}{\centering
    Insert
    \texttt{canonical\_correction\_before\_after.png}.}}
}
\caption{
Canonical one-step KRR residuals before and after the fully non-oracle
profile--spectral correction. Small and moderate extrema branches move toward
$r=0$; the largest branches remain only partly corrected.
}
\label{fig:correction_residuals}
\end{figure}

\paragraph{Recursive rollout.}

We define a rollout as recursive use of the surrogate: after the first step,
each predicted state becomes the input to the next prediction. This differs
from one-step evaluation, where each prediction starts from a true state.

Starting from the true initial state, we define the plain rollout by
\begin{equation}
\hat u^{n+1}
=
f(\hat u^n,\nu).
\label{eq:rollout_definition}
\end{equation}

To study how the error accumulates, we define the state error
\[
e_n=u_n-\hat u_n
\]
and the one-step surrogate residual evaluated at the predicted state
\begin{equation}
r_1(\hat u_n)
=
\Phi_H(\hat u_n,\nu)
-
f(\hat u_n,\nu).
\label{eq:rollout_one_step_residual}
\end{equation}

We define the Jacobian of the reference flow with respect to its state input as
\[
J_j
=
D_u\Phi_H(\hat u_j,\nu).
\]

For $0\le k\le m-1$, we define the propagation matrix
\begin{equation}
\mathcal J_{m,k}
=
\begin{cases}
J_{m-1}J_{m-2}\cdots J_{k+1},
& k<m-1,\\[2mm]
I,
& k=m-1,
\end{cases}
\label{eq:rollout_propagator}
\end{equation}
where $I$ is the identity matrix.

Assuming that $\Phi_H(\cdot,\nu)$ is twice continuously differentiable in a
neighborhood of the rollout states, we show in
Appendix~\ref{app:rollout_details}, using the standard multivariable Taylor
theorem \citep{rudin1976principles}, that the accumulated $m$-step error
satisfies, to first order,
\begin{equation}
\boxed{
e_m
\approx
\sum_{k=0}^{m-1}
\mathcal J_{m,k}\,
r_1(\hat u_k).
}
\label{eq:rollout_accumulation}
\end{equation}

Equation~\eqref{eq:rollout_accumulation} shows explicitly how one-step errors
accumulate recursively. A residual produced at step $k$ is propagated through
the later flow sensitivities $\mathcal J_{m,k}$ before contributing to the
error at step $m$. Earlier one-step errors can therefore influence all later
states.

We apply the same non-oracle correction inside a $20$-step KRR rollout. The
final relative $L^2$ error decreases from $0.935$ to $0.616$. Pooled
all-point MSE decreases from $0.0270$ to $0.0159$, and pooled near-extrema
MSE decreases from $0.0353$ to $0.0165$. On the fixed plain-rollout
active-window set, the median residual energy removed is $E=0.65$, with
$79\%$ of windows improved.

\begin{table}[H]
\centering
\small
\caption{Recursive KRR rollout over the six validation trajectories.}
\label{tab:rollout}
\begin{tabular}{lcc}
\toprule
Metric
& Plain rollout
& Corrected rollout
\\
\midrule

Final relative $L^2$
& 0.935
& 0.616
\\

Pooled all-point MSE
& 0.0270
& 0.0159
\\

Pooled near-extrema MSE
& 0.0353
& 0.0165
\\

Pooled median $|r|$
& 0.090
& 0.057
\\

Pooled $|r|\le0.05$
& 33.2\%
& 43.8\%
\\
\bottomrule
\end{tabular}
\end{table}

\begin{figure}[H]
\centering
\IfFileExists{
figures_rollout/rollout_corrected_rel_l2.png
}{
    \includegraphics[width=0.55\linewidth]
    {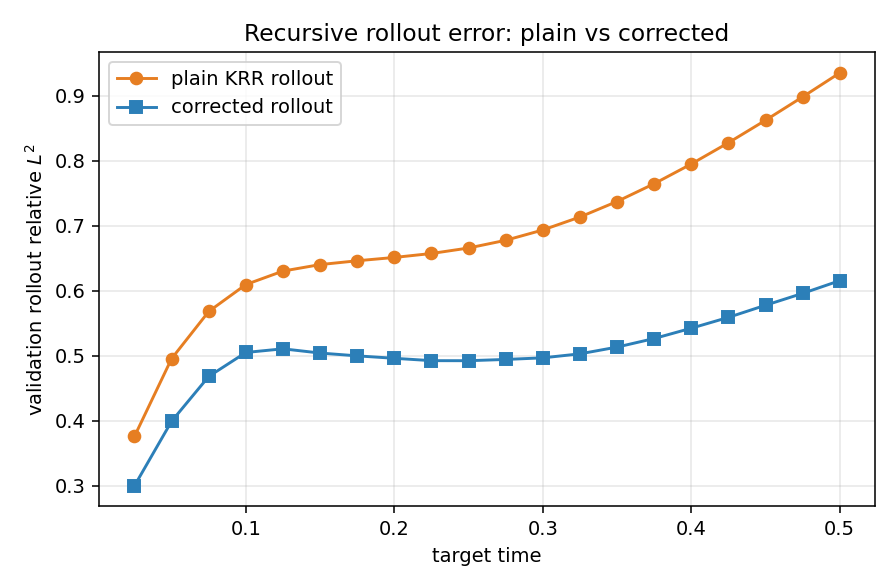}
}{
    \fbox{\parbox{0.50\linewidth}{\centering
    Insert
    \texttt{rollout\_corrected\_rel\_l2.png}.}}
}
\caption{
Relative $L^2$ error during recursive KRR rollout, with and without the
non-oracle correction applied in the loop.
}
\label{fig:rollout_curve}
\end{figure}

The correction is used as a test of whether the diagnosed local structure
contains usable information. The one-step and rollout reductions show that it
does, although the largest peak errors remain the most difficult cases.

\subsection{Discussion and limitations}
\label{sec:discussion}

The experiments support a two-level interpretation. First, extrema create a
folded value--curvature geometry that can organize residuals into curved local
branches across different estimators. Second, the physical attribution of
those branches depends on the model family. For KRR and Ridge, the clearest
moderate/high-viscosity interpretation is insufficient viscous smoothing.
The tree ensembles share the geometric pattern but show much weaker endpoint
diffusion attribution.

Several limitations are important. The study concerns one one-dimensional PDE
and one raw-grid representation, so the geometric pattern should not yet be
claimed as universal. The quantity $\nu_{\rm eff}$ is a horizon-dependent
diagnostic and should not be interpreted as a learned physical viscosity. The
diffusion regression is closely related to the earlier curvature regression,
so the negative control and spectral roughness provide important additional
evidence. Low-viscosity results are less uniform. Finally, local conic fits
generalize poorly when transferred directly to held-out windows. The main
geometric result therefore concerns the fold/extrema mechanism, while the
finite-window conics remain descriptive diagnostics.

\section{Conclusion}
\label{sec:conclusion}

Classical raw-grid surrogates for viscous Burgers can have structured local
errors that are hidden by aggregate metrics. Across KRR, Ridge, ExtraTrees,
and Random Forests, one-step residuals form coherent branches near predicted
extrema. For the KRR reference, curvature dominates transport, and combining the
local extremum fold with the curvature-based residual model predicts a
leading-order near-parabolic relation in the $(X,r)$ plane. Physics-based
tests then separate geometry from mechanism: KRR
and Ridge show evidence consistent with insufficient viscous smoothing at
moderate and high viscosity, while the tree ensembles show much weaker
diffusion attribution. Non-oracle one-step and rollout corrections further
show that the diagnosed structure carries predictive information. These
results illustrate a practical route from an observed surrogate failure to a
testable physical interpretation.


\clearpage
\appendix

\begin{center}
    {\Large\bfseries Appendix}
\end{center}
\vspace{0.5em}

\section{Full Experimental Setup}
\label{app:protocol}

We use three experimental protocols. They serve different purposes, so their
absolute errors should not be mixed.

\begin{table}[H]
\centering
\small
\caption{Summary of the three experimental protocols used in the paper.}
\label{tab:protocols}
\begin{tabular}{llll}
\toprule
Setting & Large diagnostic & Canonical & Replication \\
\midrule
Purpose
& first-order regression
& main results
& held-out conic test \\
Data
& $5{,}000$ train states,
  $20{,}000$ val. states
& $30$ trajectories
& $30$ fresh trajectories \\
Train/validation
& subsampled states
& $24/6$, seed $42$
& $24/6$ \\
Validation points
& $2{,}560{,}000$
& $15{,}360$
& $15{,}360$ \\
KRR $\alpha$
& $10^{-4}$
& $0.01$
& $0.01$ \\
RBF $\gamma$
& $10^{-4}$
& $10^{-4}$
& $10^{-4}$ \\
KRR one-step MSE
& $7.8596\times10^{-6}$
& $3.7236\times10^{-3}$
& $5.221\times10^{-3}$ \\
Relative $L^2$
& ---
& $0.2201$
& $0.2157$ \\
\bottomrule
\end{tabular}
\end{table}

The large diagnostic protocol uses a one-frame saved-time prediction horizon
from the original diagnostic data set. It is used only for the regression in
Sec.~\ref{sec:first_order_test}. Its sampling and saved-time horizon differ
from the canonical protocol, so its absolute MSE is not compared with the
canonical MSE.

The canonical protocol uses the periodic domain $[0,1)$, $N=128$, and
$\Delta x=1/128$. Reference trajectories are generated with RK4
\citep{durran2010numerical}, using solver time step $\Delta t=10^{-4}$ for
$5000$ RK4 steps. States are stored every $250$ steps, which gives the
one-step horizon $H=0.025$. Initial conditions use the five-mode family in
Eq.~\eqref{eq:initial_conditions}. One viscosity from
$\{0.01,0.02,0.05,0.1\}$ is sampled once per trajectory. The $30$
trajectories are split into $24$ training and $6$ validation trajectories
with seed $42$.

For the canonical cross-model comparison, RBF KRR uses regularization $0.01$
and RBF parameter $10^{-4}$. Linear Ridge uses penalty $1$. ExtraTrees uses
$300$ trees, feature fraction $0.7$, minimum leaf size $1$, and seed $42$.
Random Forest uses $200$ trees with the same feature fraction, minimum leaf
size, and seed.

The replication protocol repeats the canonical settings with a fresh draw of
$30$ trajectories and a fresh $24/6$ split. It is used only for the held-out
conic tests in Appendix~\ref{app:geometry_details}. Its KRR validation MSE is
$5.221\times10^{-3}$ and its relative $L^2$ error is $0.2157$.

\paragraph{Active-window definition.}

Within each extremum type separately, we define $q$ as the $90$th percentile
of $|r|$ over the radius-$3$ near-extrema points. We define a radius-$5$
window as active if its peak $|r|$ exceeds the corresponding $q$. The
active-window MSE is computed over the union of active-window grid points.

\paragraph{Local conic fit.}

Within each radius-$5$ window, we standardize $X=\hat u$ and $Y=r$ to zero
mean and unit variance. Following the standard algebraic representation of
conics \citep{fitzgibbon1999ellipse}, we form the design matrix with rows
\[
[X^2,\;XY,\;Y^2,\;X,\;Y,\;1].
\]

We define the fitted coefficient vector as the right singular vector
associated with the smallest singular value. Coefficients are normalized
before computing Eq.~\eqref{eq:conic_discriminant}.

\section{Derivations and Supporting Calculations}
\label{app:derivations}

\subsection{First-order operator perturbation}

For a sufficiently smooth periodic field, standard translation-group theory
gives \citep{engel2000semigroups}
\begin{equation}
T_\delta\hat u
=
\hat u
+
\delta\partial_x\hat u
+
O(\delta^2).
\label{eq:translation_expansion}
\end{equation}

For a sufficiently smooth field $v$, the standard heat-semigroup expansion
gives \citep{pazy1983semigroups}
\begin{equation}
e^{\beta\partial_{xx}}v
=
v
+
\beta\partial_{xx}v
+
O(\beta^2).
\label{eq:heat_expansion}
\end{equation}

We now show Eq.~\eqref{eq:first_order_residual}. Substituting
Eq.~\eqref{eq:translation_expansion} into
Eq.~\eqref{eq:operator_model}, applying
Eq.~\eqref{eq:heat_expansion}, and multiplying by $1+\alpha$ gives
\begin{align}
u_{\rm true}-\hat u
&=
\alpha\hat u
+
\delta\partial_x\hat u
+
\beta\partial_{xx}\hat u
\nonumber\\
&\quad
+
O\!\left(
\delta^2
+
\beta^2
+
|\alpha\delta|
+
|\alpha\beta|
+
|\beta\delta|
\right).
\label{eq:first_order_full}
\end{align}

Keeping only first-order terms gives
Eq.~\eqref{eq:first_order_residual}.

\subsection{Fold relation and residual relation}

We now show Eq.~\eqref{eq:fold_relation}. We define
\[
\varepsilon=x-x_0.
\]

Because $x_0$ is a differentiable interior local extremum, the standard
first-order condition gives
\[
\partial_x\hat u(x_0)=0
\]
\citep{rudin1976principles}.

Taylor's theorem \citep{rudin1976principles} gives
\begin{equation}
\hat u(x_0+\varepsilon)
=
U_0
+
\frac{c_2}{2}\varepsilon^2
+
\frac{c_3}{6}\varepsilon^3
+
O(\varepsilon^4),
\label{eq:taylor_u}
\end{equation}
and
\begin{equation}
\partial_{xx}\hat u(x_0+\varepsilon)
=
c_2
+
c_3\varepsilon
+
O(\varepsilon^2).
\label{eq:taylor_uxx}
\end{equation}

Using the definitions of $X$ and $Z$, these two expansions give
\[
X-U_0
=
\frac{c_2}{2}\varepsilon^2
+
O(\varepsilon^3),
\]
and
\[
Z-c_2
=
c_3\varepsilon
+
O(\varepsilon^2).
\]

Squaring the second relation gives
\[
(Z-c_2)^2
=
c_3^2\varepsilon^2
+
O(\varepsilon^3).
\]
Multiplying the first relation by $2c_3^2/c_2$ gives
\[
\frac{2c_3^2}{c_2}(X-U_0)
=
c_3^2\varepsilon^2
+
O(\varepsilon^3).
\]
Therefore,
\begin{equation}
(Z-c_2)^2
=
\frac{2c_3^2}{c_2}(X-U_0)
+
O(\varepsilon^3).
\label{eq:fold_with_remainder}
\end{equation}
Keeping the leading term gives Eq.~\eqref{eq:fold_relation}.

We next show Eq.~\eqref{eq:residual_conic} under the first-order local residual
model. Near the extremum, the first-derivative shift contribution is small
under the approximation used in Sec.~\ref{sec:geometric_analysis}, and
Eq.~\eqref{eq:extremum_residual} gives
\[
r
\approx
\alpha X+\beta Z.
\]
For $\beta\neq0$,
\[
Z
\approx
\frac{r-\alpha X}{\beta}.
\]
Substituting this relation into Eq.~\eqref{eq:fold_relation} gives
\[
\left(
\frac{r-\alpha X}{\beta}
-c_2
\right)^2
\approx
\frac{2c_3^2}{c_2}
(X-U_0).
\]
Multiplying by $\beta^2$ gives Eq.~\eqref{eq:residual_conic}. Thus the
$(X,r)$ relation follows from the fold relation together with the stated
first-order residual model.

If $c_3=0$, the displayed fold is not the leading asymmetric relation. In that
case, one must continue the Taylor expansion to the first nonzero odd
derivative.

\subsection{Why the residual rate is compared with Burgers terms}

We now justify the use of $r/H$ in
Eq.~\eqref{eq:physics_regression}.

From the definition of the residual, subtracting the same input state $u^n$
from the two future states gives the exact identity
\begin{equation}
\frac{r}{H}
=
\frac{u_{\rm true}-u^n}{H}
-
\frac{\hat u-u^n}{H}.
\label{eq:residual_rate_identity}
\end{equation}

For the reference solution, the fundamental theorem of calculus
\citep{rudin1976principles} gives
\begin{equation}
\frac{u(t+H)-u(t)}{H}
=
\frac{1}{H}
\int_t^{t+H}
\partial_su(s)\,ds.
\label{eq:ftc_rate}
\end{equation}

Using the Burgers equation
Eq.~\eqref{eq:burgers}, we obtain
\begin{equation}
\frac{u(t+H)-u(t)}{H}
=
\frac{1}{H}
\int_t^{t+H}
\left[
-u\partial_xu
+
\nu\partial_{xx}u
\right]ds.
\label{eq:burgers_average_rate}
\end{equation}

Equations~\eqref{eq:residual_rate_identity} and
\eqref{eq:burgers_average_rate} show that $r/H$ compares the reference and
surrogate average rates of change, while the reference rate is governed by
the Burgers advection and diffusion terms. This provides the stated
motivation for the endpoint features $A(\hat u)$ and $D(\hat u,\nu)$ in the
diagnostic regression.

\subsection{Effective-diffusion diagnostic}

We now show Eq.~\eqref{eq:effective_diffusion_ratio} under the local diffusion
model of Eq.~\eqref{eq:effective_diffusion_model}. In this diagnostic,
$u_{\rm sur}$ denotes the surrogate endpoint $\hat u$.

For sufficiently regular $v$, the standard heat-semigroup expansion gives
\citep{pazy1983semigroups}
\[
e^{H\nu\partial_{xx}}v
=
v
+
H\nu\partial_{xx}v
+
O(H^2),
\]
and
\[
e^{H\nu_{\rm eff}\partial_{xx}}v
=
v
+
H\nu_{\rm eff}\partial_{xx}v
+
O(H^2).
\]
Subtracting gives
\[
u_{\rm true}-u_{\rm sur}
=
H(\nu-\nu_{\rm eff})\partial_{xx}v
+
O(H^2).
\]
Dividing by $H$ gives
\begin{equation}
\frac{u_{\rm true}-u_{\rm sur}}{H}
=
(\nu-\nu_{\rm eff})\partial_{xx}v
+
O(H).
\label{eq:effective_diffusion_expansion}
\end{equation}

To compare this expression with the regression feature
$\partial_{xx}\hat u$, we make one additional short-horizon regularity
assumption: $v$ is smooth enough that the heat-semigroup expansion remains
valid after applying two spatial derivatives. Since $u_{\rm sur}=\hat u$, the
same expansion then gives
\[
\partial_{xx}\hat u
=
\partial_{xx}v
+
O(H).
\]
Hence
\[
\partial_{xx}v
=
\partial_{xx}\hat u
+
O(H),
\]
and Eq.~\eqref{eq:effective_diffusion_expansion} becomes
\[
\frac{u_{\rm true}-\hat u}{H}
=
(\nu-\nu_{\rm eff})\partial_{xx}\hat u
+
O(H).
\]

The diffusion part of Eq.~\eqref{eq:physics_regression} is
$\gamma\nu\partial_{xx}\hat u$. Matching the leading coefficient of
$\partial_{xx}\hat u$ gives
\[
\gamma\nu
\approx
\nu-\nu_{\rm eff}.
\]
Since $\nu>0$,
\[
\frac{\nu_{\rm eff}}{\nu}
\approx
1-\gamma,
\]
which is Eq.~\eqref{eq:effective_diffusion_ratio}. This is a leading-order
short-horizon diagnostic relation under the assumptions stated above.

\subsection{Why the roughness diagnostic uses fourth-power frequency weighting}

We now justify the $k^4$ weighting in
Eq.~\eqref{eq:roughness}.

For a periodic Fourier series,
\[
u(x)
=
\sum_k
\widetilde u_k
e^{2\pi ikx/L}.
\]

Standard Fourier differentiation gives
\citep{trefethen2000spectral}
\[
\partial_{xx}u
=
-
\sum_k
\left(
\frac{2\pi k}{L}
\right)^2
\widetilde u_k
e^{2\pi ikx/L}.
\]

Applying Parseval's identity gives
\begin{equation}
\|\partial_{xx}u\|_2^2
=
\sum_k
\left(
\frac{2\pi k}{L}
\right)^4
|\widetilde u_k|^2.
\label{eq:parseval_curvature}
\end{equation}

Equation~\eqref{eq:parseval_curvature} shows that second-derivative energy
weights Fourier mode $k$ by $k^4$, up to the common constant
$(2\pi/L)^4$. Normalizing by the total Fourier energy
$\sum_k|\widetilde u_k|^2$ gives the roughness measure defined in
Eq.~\eqref{eq:roughness}.

\section{Additional Geometric Results}
\label{app:geometry_details}

\subsection{Finite-window conic classes}

Using the conic classification defined in
Sec.~\ref{sec:problem_method}, we obtain the counts in
Table~\ref{tab:conic_counts}.

\begin{table}[H]
\centering
\small
\caption{Canonical active-window conic classifications.}
\label{tab:conic_counts}
\begin{tabular}{lrrrr}
\toprule
Type
& Ellipse-like
& Hyperbola-like
& Parabola-like
& Total
\\
\midrule

Maxima
& 24
& 29
& 0
& 53
\\

Minima
& 26
& 13
& 1
& 40
\\
\bottomrule
\end{tabular}
\end{table}

These empirical classes contain finite-window and higher-order effects. The
leading-order relation shown in Eq.~\eqref{eq:residual_conic} is parabolic.

\begin{figure}[H]
\centering
\IfFileExists{
figures_part1_conditions/stepF_implicit_conic_representative_arcs.png
}{
    \includegraphics[width=0.80\linewidth]
    {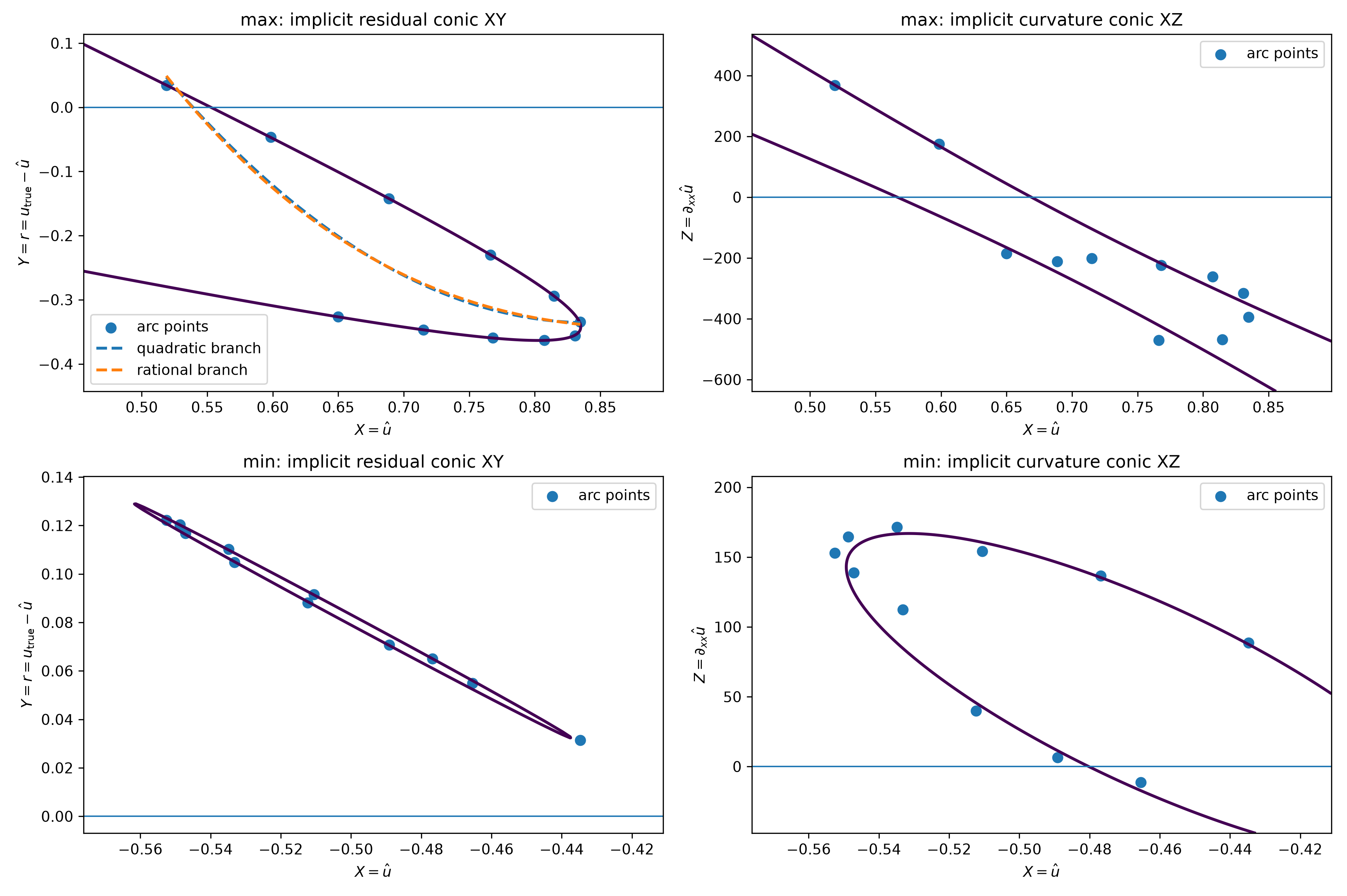}
}{
    \fbox{\parbox{0.74\linewidth}{\centering
    Insert
    \texttt{stepF\_implicit\_conic\_representative\_arcs.png}.}}
}
\caption{
Representative KRR windows and their fitted implicit conics in the $(X,r)$
and $(X,Z)$ planes. Some fitted conics follow the observed local branch,
while others select a different conic type from the shape suggested by the
data. These examples illustrate both the local geometric structure and the
limits of finite-window conic fitting.
}
\label{fig:representative_conics}
\end{figure}

\begin{figure}[H]
\centering
\IfFileExists{
figures_part1_conditions/stepN_discriminant_histograms_active_arcs.png
}{
    \includegraphics[width=0.78\linewidth]
    {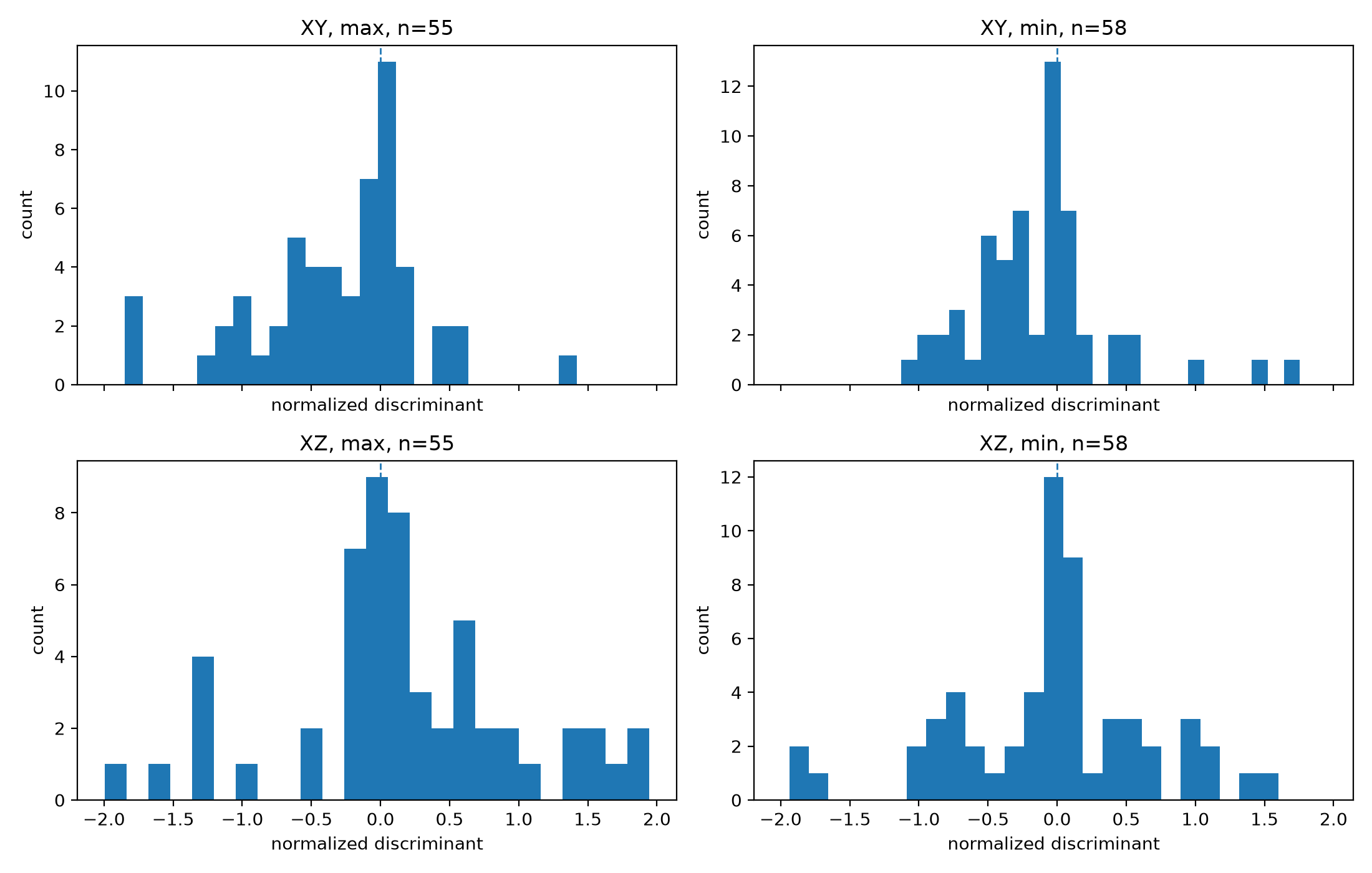}
}{
    \fbox{\parbox{0.72\linewidth}{\centering
    Insert
    \texttt{stepN\_discriminant\_histograms\_active\_arcs.png}.}}
}
\caption{
Distributions of normalized finite-window conic discriminants for active
KRR windows.
}
\label{fig:discriminant_hist}
\end{figure}

\subsection{Held-out conic transfer}

To test whether a fitted local conic is a reusable prediction rule, we use
the independent replication protocol defined in Appendix~\ref{app:protocol}.
For each active $11$-point window, the within-window test fits the conic on
$6$ points and evaluates it on the remaining $5$ points. This is our held-out
conic test; it is separate from the canonical experiment.

\begin{table}[H]
\centering
\small
\caption{
Within-window held-out conic RMSE on the independent replication protocol.
The conic is fit on $6$ of the $11$ points and evaluated on the remaining $5$.
The large held-out gap shows that local conics are useful as descriptive
diagnostics but transfer poorly as prediction rules.
}
\label{tab:heldout_conics}
\begin{tabular}{llccc}
\toprule
Extremum
& Class
& All
& Train
& Held-out
\\
\midrule

Max
& ellipse-like
& 0.038
& 0.006
& 0.145
\\

Max
& hyperbola-like
& 0.036
& 0.007
& 0.152
\\

Min
& ellipse-like
& 0.039
& 0.009
& 0.154
\\

Min
& hyperbola-like
& 0.017
& 0.005
& 0.107
\\
\bottomrule
\end{tabular}
\end{table}

\section{Full Physical Sweep}
\label{app:physics_sweep}

For completeness, we also fit a single physics regression pooled across all
viscosities. This gives
\[
\gamma_{\rm KRR}
=
0.3157,
\qquad
95\%~{\rm CI}
=
[0.2765,0.3383],
\]
and
\[
\gamma_{\rm Ridge}
=
0.3407,
\qquad
95\%~{\rm CI}
=
[0.3170,0.3613].
\]

These are coefficients from one regression fitted jointly across all
viscosities. They are not averages of the coefficients, or of
$\nu_{\rm eff}/\nu$, obtained from the separate per-viscosity regressions.
For this reason, the pooled effective-diffusion value can lie outside the
range of the per-viscosity values. We use the separate moderate- and
high-viscosity results for the physical interpretation in the main text.

At $\nu=0.05$ and $H=0.025$, the fitted values are
\[
\gamma_{\rm KRR}
=
0.2720,
\qquad
95\%~{\rm CI}
=
[0.2190,0.3217],
\]
and
\[
\gamma_{\rm Ridge}
=
0.3190,
\qquad
95\%~{\rm CI}
=
[0.2409,0.3893].
\]

At $\nu=0.10$ and $H=0.025$, the fitted values are
\[
\gamma_{\rm KRR}
=
0.2497,
\qquad
95\%~{\rm CI}
=
[0.2196,0.2820],
\]
and
\[
\gamma_{\rm Ridge}
=
0.2884,
\qquad
95\%~{\rm CI}
=
[0.2484,0.3196].
\]

Using Eq.~\eqref{eq:effective_diffusion_ratio}, we obtain the full
effective-diffusion values reported in Table~\ref{tab:full_nueff}.

\begin{table}[H]
\centering
\small
\caption{
Full endpoint effective-diffusion diagnostic.
Entries are $\nu_{\rm eff}/\nu$ from separate per-viscosity regressions.
}
\label{tab:full_nueff}
\begin{tabular}{llrrrr}
\toprule
Model
& $\nu$
& $H=.025$
& $H=.05$
& $H=.10$
& $H=.20$
\\
\midrule

KRR & 0.01 & 1.267 & 0.689 & 0.476 & 0.570 \\
KRR & 0.02 & 1.094 & 0.938 & 0.788 & 0.731 \\
KRR & 0.05 & 0.728 & 0.765 & 0.848 & 0.920 \\
KRR & 0.10 & 0.750 & 0.860 & 0.932 & 0.965 \\

Ridge & 0.01 & 0.459 & 0.415 & 0.590 & 0.763 \\
Ridge & 0.02 & 0.817 & 0.791 & 0.714 & 0.746 \\
Ridge & 0.05 & 0.681 & 0.734 & 0.813 & 0.911 \\
Ridge & 0.10 & 0.712 & 0.854 & 0.933 & 0.958 \\
\bottomrule
\end{tabular}
\end{table}

The low-viscosity rows are less coherent. In particular, KRR gives
$\nu_{\rm eff}/\nu>1$ at $\nu=0.01$ and $H=0.025$. We therefore restrict the
insufficient-smoothing interpretation to the moderate- and high-viscosity
regimes.

For reference, the corresponding curvature-energy ratios
(prediction divided by true target) at $H=0.025$ are
\[
\begin{array}{c|cc}
& \nu=0.05 & \nu=0.10\\
\hline
{\rm KRR} & 2.3494 & 3.3894\\
{\rm Ridge} & 3.1304 & 5.8239
\end{array}
\]
and at $H=0.20$ are
\[
\begin{array}{c|cc}
& \nu=0.05 & \nu=0.10\\
\hline
{\rm KRR} & 6.3002 & 17.1209\\
{\rm Ridge} & 9.6401 & 19.9307
\end{array}.
\]

\section{Correction Details and Ablations}
\label{app:correction_details}

For an active local window with true residual $r_i$ and predicted residual
correction $\hat r_i$, we define the fraction of residual energy removed as
\begin{equation}
E
=
1-
\frac{
\sum_i
(r_i-\hat r_i)^2
}{
\sum_i
r_i^2
}.
\label{eq:energy_removed}
\end{equation}

By this definition, $E=1$ means that all residual energy in the window is
removed, while $E=0$ means that the squared residual energy is unchanged.

The fully non-oracle correction uses sign-canonicalized radius-$5$ profiles of
$\hat u$, $\partial_x\hat u$, and $\partial_{xx}\hat u$, meaning that the two
extremum types are represented in comparable coordinates before fitting. The
features also include local scalar summaries and $\nu$. Maxima and minima are
modeled separately. The
main profile predictor uses gradient boosting \citep{friedman2001gbm}.
Overlapping predicted windows are combined with center-weighted triangular
aggregation, which gives larger weight to points closer to the window center.
The spectral calibration learns frequency-band gains using training data
only. The PCA-template baseline uses standard low-dimensional
principal-component ideas \citep{jolliffe2002pca,berkooz1993pod}.

\begin{table}[H]
\centering
\small
\caption{
Full non-oracle one-step KRR correction ablation.
Reductions are relative to plain KRR.
}
\label{tab:correction_ablation}
\begin{tabular}{lccccc}
\toprule
Method
& All red.
& Near red.
& Active red.
& Rel.\ $L^2$
& Median $E$
\\
\midrule

Profile correction
& 24.29\%
& 43.99\%
& 39.80\%
& 0.192
& 0.448
\\

Profile + spectral filter
& 31.24\%
& 49.54\%
& 45.22\%
& 0.183
& 0.534
\\

PCA-template
& 12.36\%
& 20.69\%
& 22.26\%
& 0.206
& 0.353
\\

RF local-window squeeze
& 8.48\%
& 11.22\%
& 29.51\%
& 0.211
& 0.410
\\

Tail-aware ensemble
& 31.08\%
& 49.51\%
& 45.42\%
& 0.183
& 0.531
\\
\bottomrule
\end{tabular}
\end{table}

The profile--spectral method is the main correction because the larger
tail-aware ensemble gives almost the same result.

\paragraph{Qualitative corrected-state snapshot.}

\begin{figure}[H]
\centering

\IfFileExists{
figures_correction/truevspredvscorr_one_step.png
}{
    \includegraphics[width=0.68\linewidth]
    {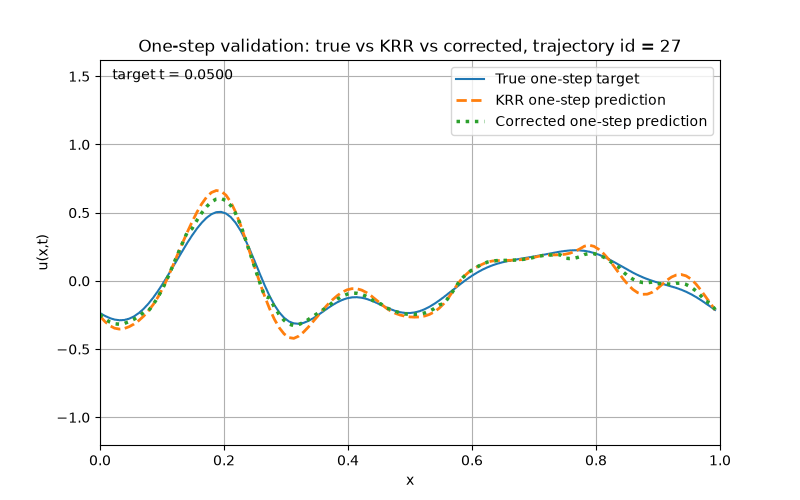}
}{
  \IfFileExists{
  truevspredvscorr_one_step.png
  }{
    \includegraphics[width=0.68\linewidth]
    {truevspredvscorr_one_step.png}
  }{
    \fbox{\parbox{0.62\linewidth}{\centering
    Insert \texttt{truevspredvscorr\_one\_step.png}.}}
  }
}

\caption{
One-step true target, plain KRR prediction, and corrected prediction for one
validation state. The correction brings several smaller extrema closer to the
reference, while some of the largest extrema remain difficult and can be
moved too strongly. This figure is qualitative; the correction metrics are
computed over the full canonical validation set.
}
\label{fig:corrected_state_appendix}
\end{figure}

Figure~\ref{fig:corrected_state_appendix} gives a direct view of the
one-step prediction before and after correction. It is kept in the appendix
because the main quantitative evidence is already given by the residual plot
and error tables.

\section{Recursive Rollout Details}
\label{app:rollout_details}

We now show Eq.~\eqref{eq:rollout_accumulation}.

By the definitions of the reference and surrogate rollouts,
\[
u_{n+1}
=
\Phi_H(u_n,\nu),
\qquad
\hat u_{n+1}
=
f(\hat u_n,\nu).
\]

Using the definition
\[
e_n=u_n-\hat u_n
\]
and the one-step residual defined in
Eq.~\eqref{eq:rollout_one_step_residual}, we obtain the exact identity
\begin{align}
e_{n+1}
&=
\Phi_H(u_n,\nu)
-
f(\hat u_n,\nu)
\nonumber\\
&=
\Phi_H(u_n,\nu)
-
\Phi_H(\hat u_n,\nu)
+
r_1(\hat u_n).
\label{eq:rollout_exact_identity}
\end{align}

Because
\[
u_n=\hat u_n+e_n,
\]
we assume that $\Phi_H(\cdot,\nu)$ is twice continuously differentiable in
a neighborhood of the rollout states. The standard multivariable Taylor
theorem \citep{rudin1976principles} then gives
\begin{equation}
\Phi_H(u_n,\nu)
=
\Phi_H(\hat u_n,\nu)
+
J_ne_n
+
R_n,
\qquad
\|R_n\|_2=O(\|e_n\|_2^2),
\label{eq:rollout_taylor}
\end{equation}
where, by definition,
\[
J_n
=
D_u\Phi_H(\hat u_n,\nu).
\]

Substituting Eq.~\eqref{eq:rollout_taylor} into
Eq.~\eqref{eq:rollout_exact_identity} gives
\begin{equation}
e_{n+1}
=
J_ne_n
+
r_1(\hat u_n)
+
R_n,
\qquad
\|R_n\|_2=O(\|e_n\|_2^2).
\label{eq:rollout_one_step_recursion}
\end{equation}

Keeping only the first-order terms in the rollout error gives
\[
e_{n+1}
\approx
J_ne_n+r_1(\hat u_n).
\]

Because the rollout starts from the true initial state,
\[
e_0=0.
\]

Repeated substitution gives
\[
e_1
\approx
r_1(\hat u_0),
\]
\[
e_2
\approx
J_1r_1(\hat u_0)
+
r_1(\hat u_1),
\]
and
\[
e_3
\approx
J_2J_1r_1(\hat u_0)
+
J_2r_1(\hat u_1)
+
r_1(\hat u_2).
\]

The formulas below keep only these first-order terms; the propagated
second-order remainders are omitted. We now show the general first-order form
by induction. Assume that
\[
e_m
\approx
\sum_{k=0}^{m-1}
\mathcal J_{m,k}r_1(\hat u_k).
\]

Using the first-order recursion gives
\[
e_{m+1}
\approx
J_me_m+r_1(\hat u_m).
\]

Substituting the induction hypothesis gives
\[
e_{m+1}
\approx
\sum_{k=0}^{m-1}
J_m\mathcal J_{m,k}r_1(\hat u_k)
+
r_1(\hat u_m).
\]

By the definition of the propagation matrices,
\[
J_m\mathcal J_{m,k}
=
\mathcal J_{m+1,k},
\]
and
\[
\mathcal J_{m+1,m}=I.
\]

Therefore,
\[
e_{m+1}
\approx
\sum_{k=0}^{m}
\mathcal J_{m+1,k}r_1(\hat u_k).
\]

This shows by induction that
\[
e_m
\approx
\sum_{k=0}^{m-1}
\mathcal J_{m,k}r_1(\hat u_k),
\]
which is Eq.~\eqref{eq:rollout_accumulation}.

\begin{table}[H]
\centering
\small
\caption{Additional recursive KRR rollout statistics.}
\label{tab:rollout_appendix}
\begin{tabular}{lccc}
\toprule
& One-step
& Plain rollout
& Corrected rollout
\\
\midrule

Final rel.\ $L^2$
& ---
& 0.935
& 0.616
\\

Pooled all-point MSE
& ---
& 0.0270
& 0.0159
\\

Pooled near-extrema MSE
& ---
& 0.0353
& 0.0165
\\

Pooled median $|r|$
& 0.011
& 0.090
& 0.057
\\

Pooled $|r|\le0.05$
& 87.7\%
& 33.2\%
& 43.8\%
\\

Active-window median $E$
& ---
& \multicolumn{2}{c}{0.65 (79\% improved)}
\\
\bottomrule
\end{tabular}
\end{table}

\begin{figure}[H]
\centering
\IfFileExists{
figures_rollout/fair_residual_scatter_one_step_vs_plain_rollout_pooled.png
}{
    \includegraphics[width=0.82\linewidth]
    {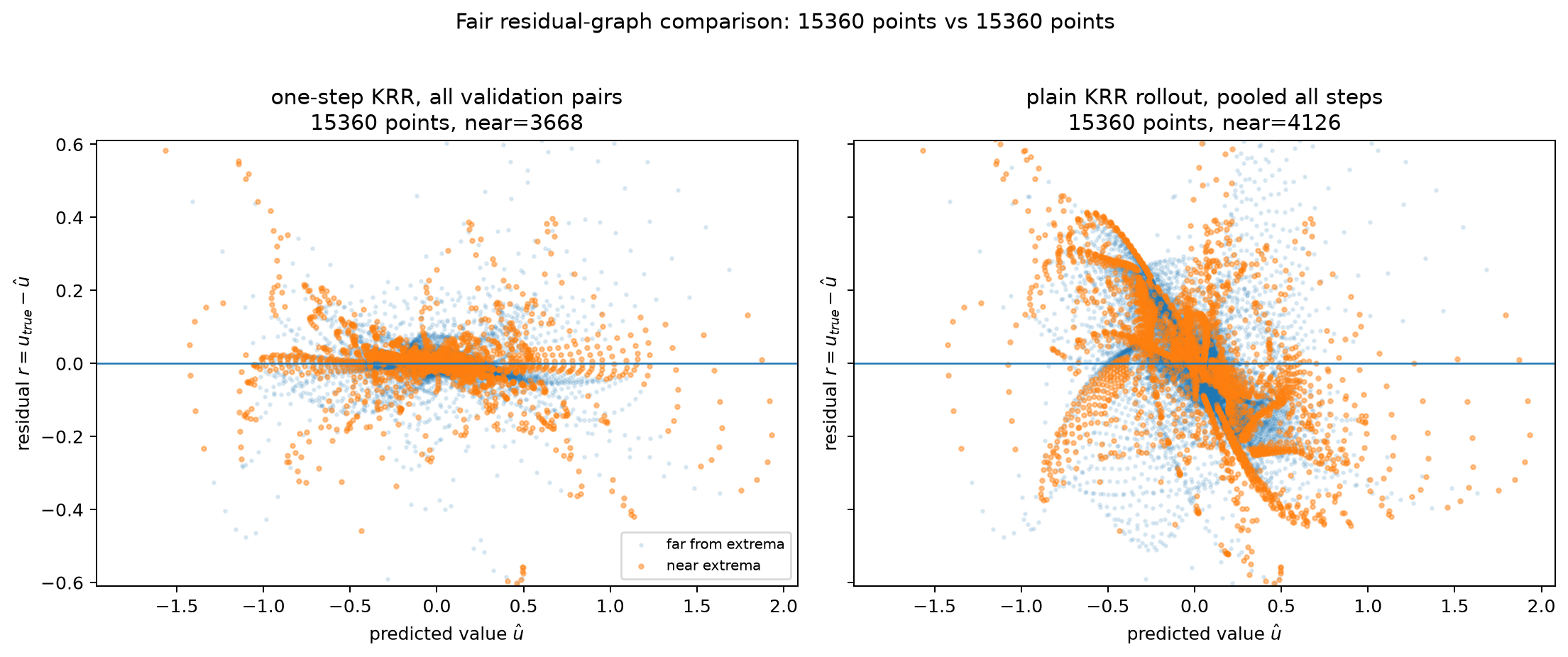}
}{
    \fbox{\parbox{0.76\linewidth}{\centering
    Insert
    \texttt{fair\_residual\_scatter\_one\_step\_vs\_plain\_rollout\_pooled.png}.}}
}
\caption{
Matched-sample comparison of one-step and recursive KRR residuals. The
near-zero cloud becomes less dominant during rollout, while the structured
off-axis branches remain visible.
}
\label{fig:rollout_residuals}
\end{figure}

\begin{figure}[H]
\centering
\IfFileExists{
figures_rollout/rollout_active_arcs_plain_vs_corrected.png
}{
    \includegraphics[width=0.82\linewidth]
    {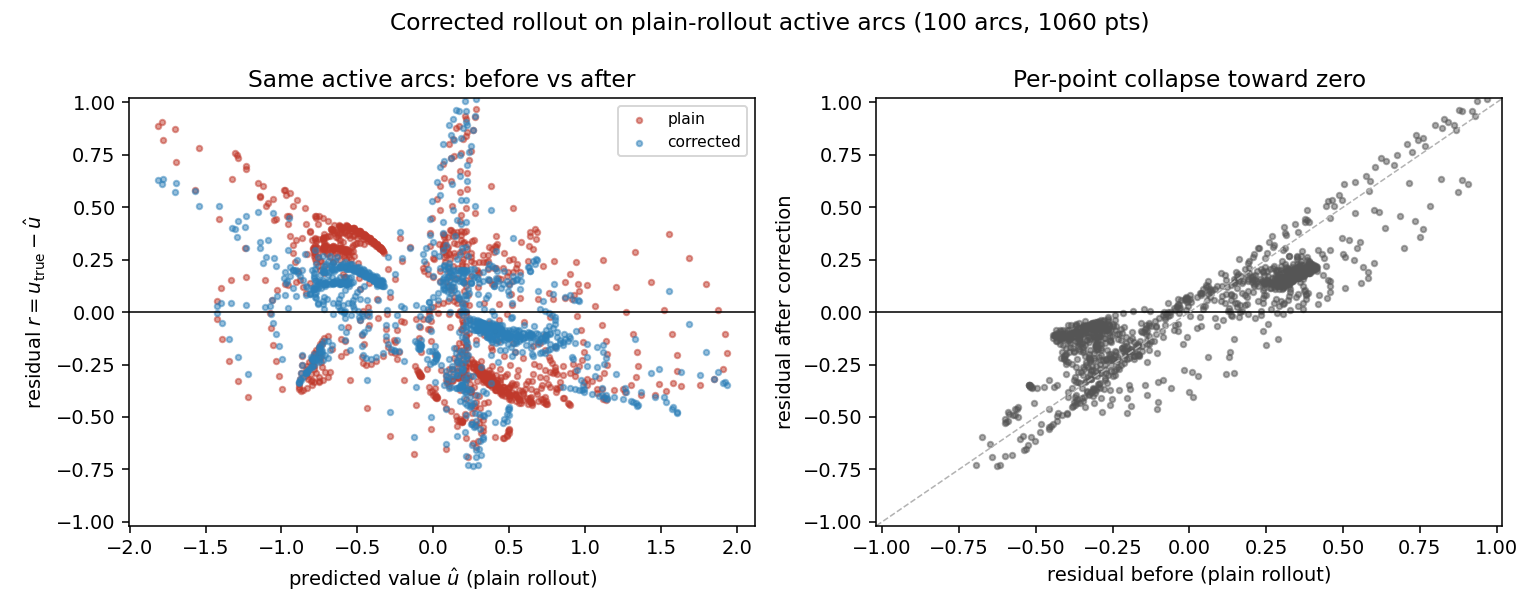}
}{
    \fbox{\parbox{0.76\linewidth}{\centering
    Insert
    \texttt{rollout\_active\_arcs\_plain\_vs\_corrected.png}.}}
}
\caption{
Plain versus corrected residuals on the fixed active-window set during
recursive rollout.
}
\label{fig:rollout_active}
\end{figure}

\section{Additional Robustness Checks}
\label{app:robustness}

The cross-model conclusion is not driven by one random tree seed. Across three
seeds, the measured number of active windows is
\[
67.3\pm2.1
\]
for ExtraTrees and
\[
66.7\pm1.5
\]
for Random Forests, where the uncertainty is the standard deviation across
the three seeds.

\begin{figure}[H]
\centering
\IfFileExists{
figures/error_graph_krr_no_gain.png
}{
    \includegraphics[width=0.74\linewidth]
    {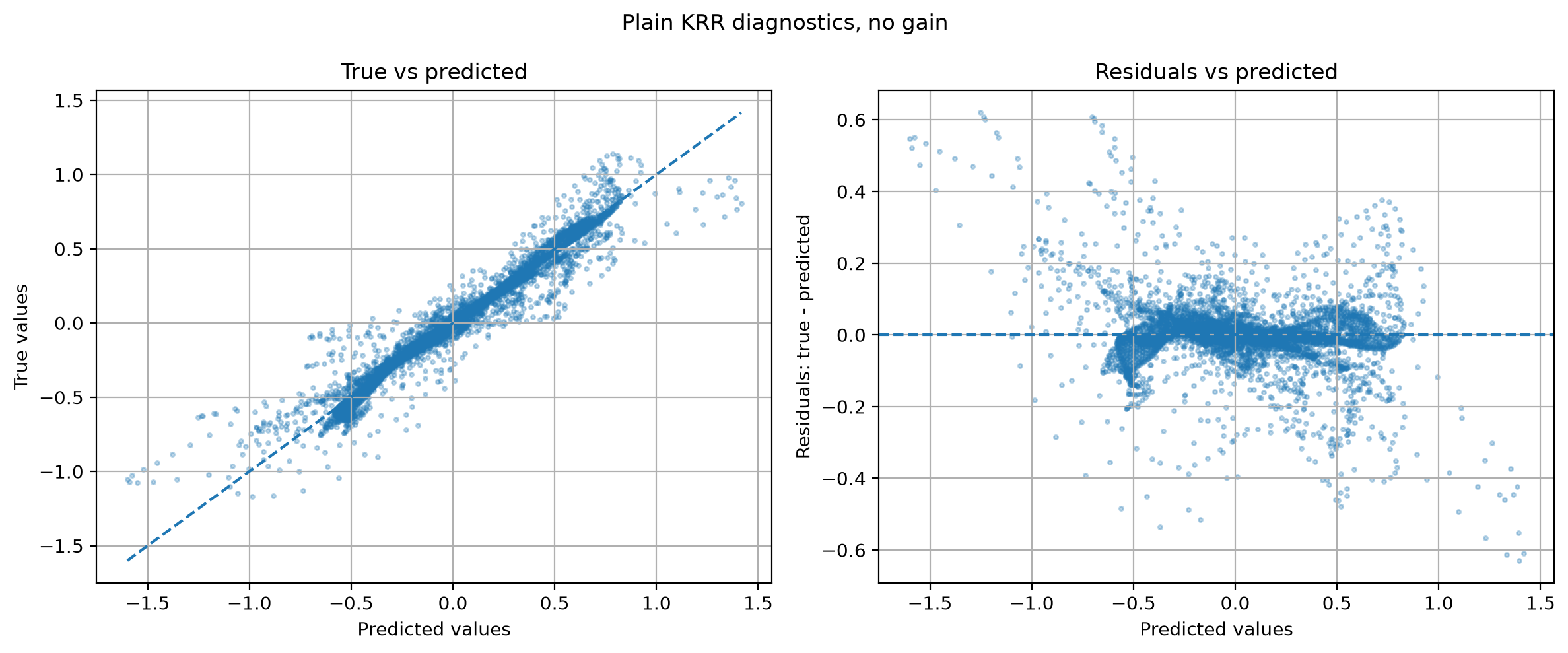}
}{
    \fbox{\parbox{0.68\linewidth}{\centering
    Insert
    \texttt{figures/error\_graph\_krr\_no\_gain.png}.}}
}
\caption{
Plain KRR residual diagnostic without rollout gain. The structured off-axis
branches remain visible.
}
\label{fig:krr_plain}
\end{figure}

For completeness, Fig.~\ref{fig:krr_plain} shows the plain KRR residual
diagnostic without any rollout-calibrated gain.

Coverage-preserving trimmed conic refits change the measured finite-window
conic class for only about $6$--$8\%$ of active windows. The held-out transfer
results in Table~\ref{tab:heldout_conics} show that the conic fits should
remain descriptive local diagnostics.


\end{document}